\pdfoutput=1

\documentclass[11pt]{article}

\usepackage[final]{acl}

\usepackage{times}
\usepackage{latexsym}

\usepackage[T1]{fontenc}

\usepackage[utf8]{inputenc}

\usepackage{microtype}

\usepackage{inconsolata}

\usepackage{graphicx}

\usepackage{booktabs}  
\usepackage{amsmath}
\usepackage{amssymb}

\usepackage{afterpage}
\usepackage{xcolor}    
\usepackage{tcolorbox} 
\usepackage{soul}
\sethlcolor{red!30} 
\usepackage{fontawesome5}
\usepackage{float}

\title{Multi-lingual Functional Evaluation for Large Language Models}

\author{
    Victor Ojewale$^{1}$, \quad
    Inioluwa Deborah Raji$^{2}$, \quad
    Suresh Venkatasubramanian$^{1}$ \\
    \\
    $^{1}$The Center for Tech Responsibility, Brown University, USA \\
    $^{2}$University of California, Berkeley, USA \\
    \\
}

\begin{document}
\maketitle
\begin{abstract}

Multi-lingual competence in large language models is often evaluated via static data benchmarks such as Belebele, M-MMLU and M-GSM. However, these evaluations often fail to provide an adequate understanding of the practical performance and robustness of models across multi-lingual settings.
In response, we create multi-lingual \emph{functional} benchmarks -- Cross-Lingual Grade School Math Symbolic (CL-GSM Symbolic) and Cross-Lingual Instruction-Following Eval (CL-IFEval)-- by translating existing functional benchmark templates from English to five additional languages that span the range of resources available for NLP: French, Spanish, Hindi, Arabic and Yoruba. 
Our results show that the gap between static and functional evaluations is highly uneven: across models, performance drops from M-GSM to CL-GSM Symbolic by 24\%, 17\%, and 18\% in English, French, and Spanish, while the drop from Belebele to CL-IFEval ranges from 15\% to 24\% across languages, and the drop from M-MMLU to CL-IFEval is much smaller (0.5\% to 3\%).Similarly, we find that model \emph{robustness} across languages varies significantly, with certain languages (eg. Arabic, English) being the most consistently well performing across evaluation iterations.

\end{abstract}

\section{Introduction}

\begin{figure*}[t]
    \centering
 \includegraphics[height=2in]{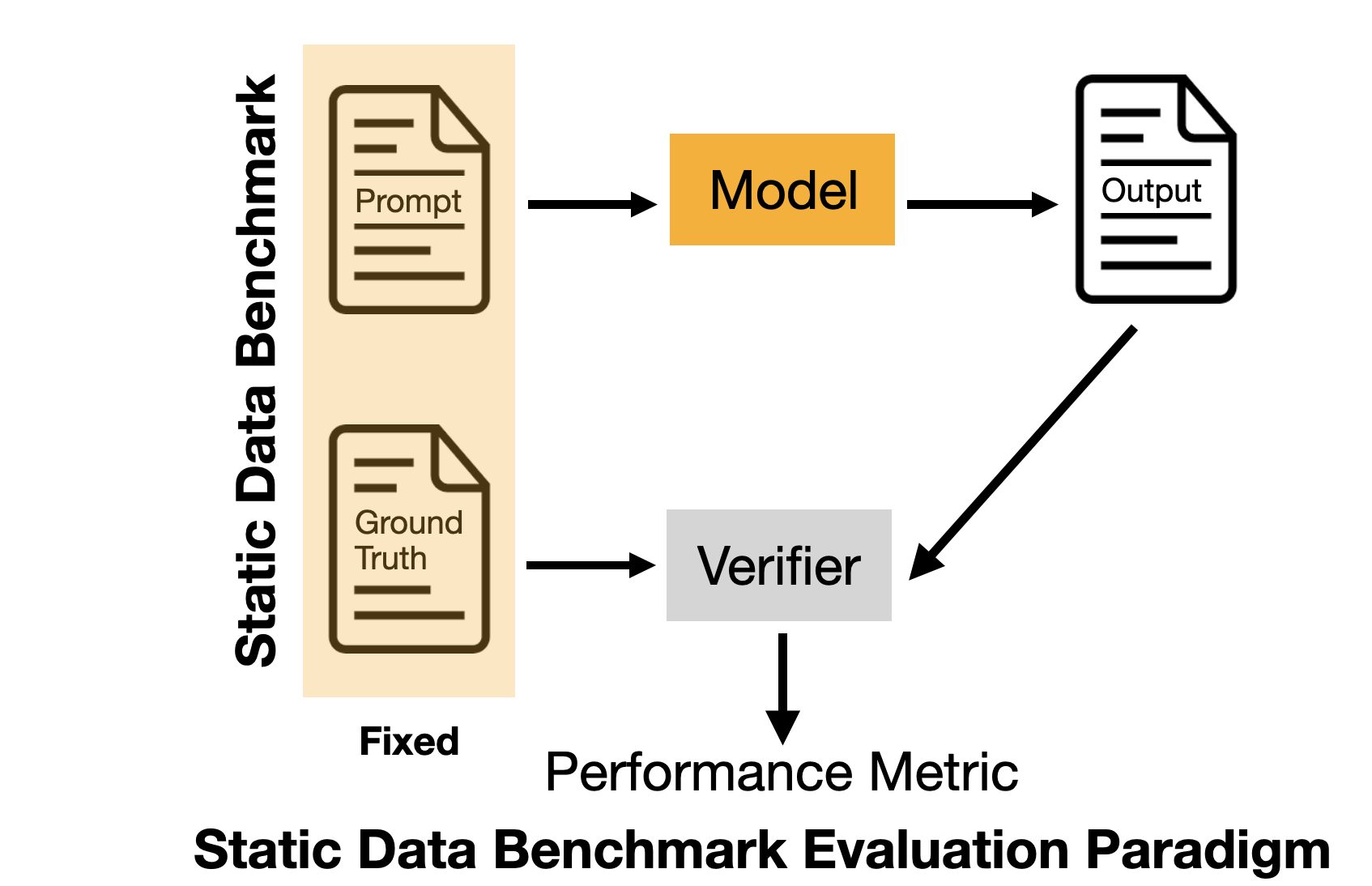}
 \includegraphics[height=2in]{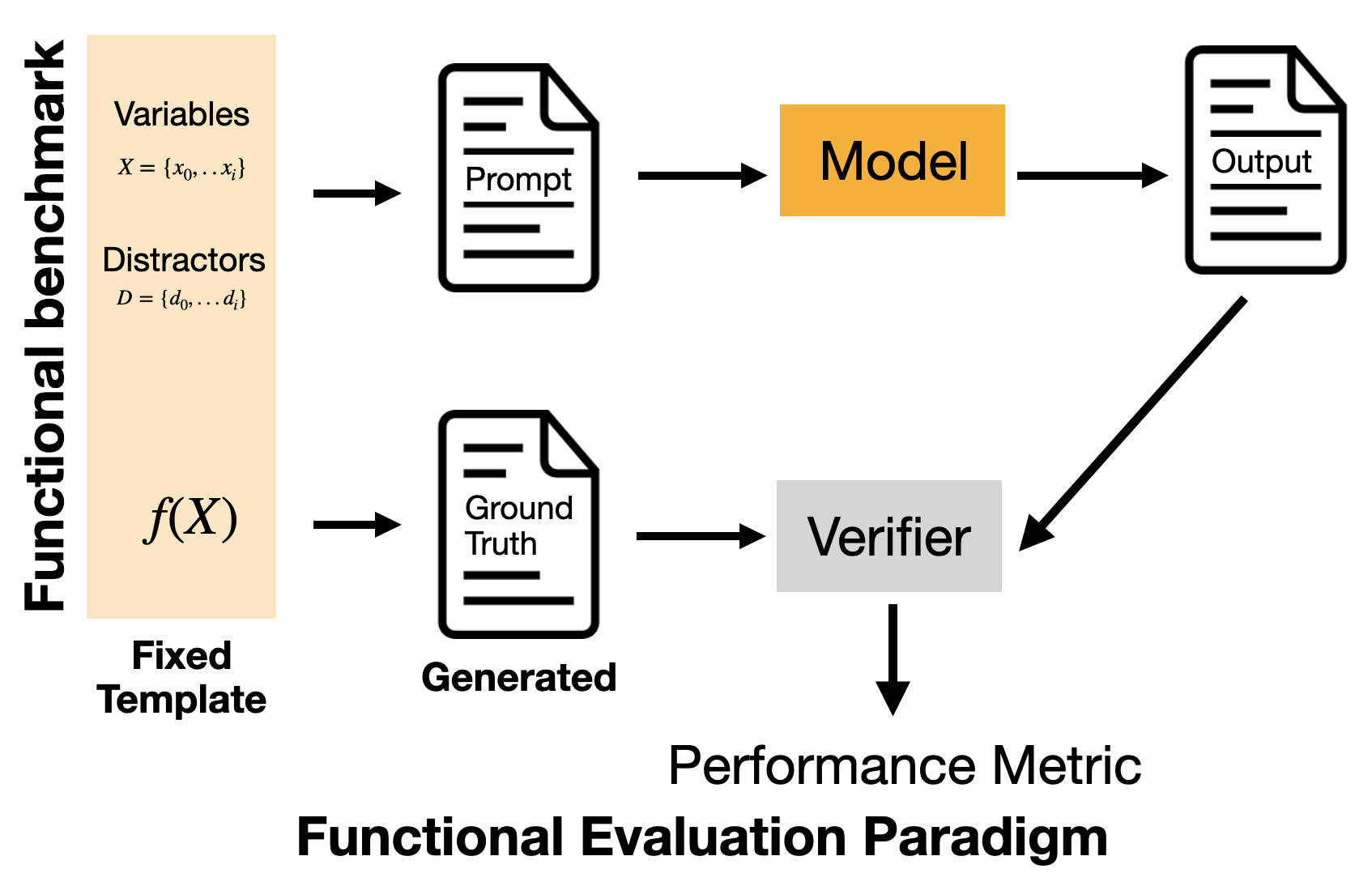}
    \caption{\textbf{Description of the functional evaluation paradigm.} Unlike with static data benchmarks, in the functional evaluation paradigm, model input prompts are not fixed but generated through a fixed template and a set of variables $X$ (modifiable prompt attributes meant to impact model outputs) and a set of distractors $D$ (modifiable prompt attributes meant to be ignored). The ground truth in this setting is generated through a fixed functional transformation $f(X)$. For instance, the prompt "Sally bought 2 red apples and 3 green apples. How much fruit did Sally buy?" is generated from the fixed template "$\{name\}$ bought $\{n_1\}$ $\{color_1\}$ apples and $\{n_2\}$ $\{color_2\}$ apples. How much fruit did $\{name\}$ buy?". This template involves the variables $X =\{n_1, n_2\}$ and the distractors $ D =\{name, color_1, color_2\}$. The correct fixed output function in this case is $f(X)=n_1+n_2$. } 
    \label{fig:overview}
\end{figure*}

Despite some meaningful progress, models operating in languages other than English have been regularly found to be more biased~\cite{talat2022you}, less safe~\cite{yong2023low} and overall meaningfully less performant and robust ~\cite{ojo2023good}. 

Popular multi-lingual LLM evaluations, such as Multilingual MMLU (M-MMLU) and the Multilingual Grade School Math (M-GSM) benchmark\citep{hendrycks2020measuring, shi2023mgsm,cobbe2021training}, 
while useful, often fail to capture more meaningful indications of \emph{functional} multi-lingual model performance -- that is, the robust execution of a given prompt across a variety of languages (see Figure ~\ref{fig:overview}).
In this paper, we extend the scope of two English "functional" evaluation datasets -- IFEval~\cite{zhou2023ifeval}, and GSM-Symbolic~\cite{gsm-symbolic} -- by translating its prompt templates into five additional languages: French, Spanish, Hindi, Arabic and Yoruba. 

Our experiments show that the relationship between functional and static benchmark performance varies significantly across models, languages, and benchmarks. While models often score higher on static benchmarks than on functional evaluations, the size of this gap varies substantially, with some static benchmarks aligning much more closely with functional performance than others. Language performance gaps also vary significantly across settings as their magnitude and direction shift across specific languages and models. Finally, functional benchmarks surface robustness that varies significantly by language and by prompt structure, revealing that models can be stable for certain languages yet brittle for others on particular instruction types, templates, or question families.

\section{Related Work}

Common multi-lingual data benchmarks such as M-MMLU \citep{hendrycks2020measuring, dac2023okapi}, FLORES \citep{goyal2022flores}, BeleBele \citep{bandarkar-etal-2024-belebele}, and XLSum \citep{hasan2021xlsum} 
possess known limitations. The use of direct translation for many of these benchmarks has been critiqued by some as being devoid of realistic cultural context~\cite{romanou2024include, singh2024global}. Furthermore, research reveals that English language benchmark data contamination might distort reported benchmark performance in English or possibly additional languages (as is the case for MMLU~\cite{dodge2021documenting} and GSM~\cite{zhang2024careful}). 

Functional evaluation involves "templating" a common popular benchmark with modifiable variables. For example, GSM-Symbolic templates examples of the static data benchmark GSM8k ~\cite{cobbe2021training} to generate the input permutations. The ground truth for the math problems is then calculated  using literal template-based functional mappings from input values to the expected output (see Figure ~\ref{fig:overview}). Recent work has attempted to set up similar symbolic annotations for natural language benchmarks~\cite{hennigen2023towards}, and we can see a similar verifiable, function-based template format with instruction-following benchmarks \citep{liu2024mitigatinghallucinationlargemultimodal,chang2023survey} such as the IFEval dataset\citet{zhou2023ifeval}. Although some concurrent work -- a proposed Multi-lingual IFEval (M-IFEval) ~\cite{dussolle2025m} -- has attempted to translate the IFEval template to Spanish, French and Japanese, we have yet to see more systematic analysis of how such functional evaluations can inform better assessments of performance and robustness in multi-lingual deployment settings.

\begin{table*}[ht]
    \centering 
    \resizebox{\linewidth}{!}{
    \begin{tabular}{lccccc}
        \toprule
        Benchmark & Avail. Lang & Prompts/Lang & Templates & Samples/Template & Task \\
        \midrule
        \multicolumn{6}{c}{\textbf{Functional Benchmarks}}\\
        \midrule
        Cross-Lingual GSM Symbolic (Ours)      & 5 & 5000 & 100 & 50 & Math Reasoning  \\
        Cross-Lingual IFEval (Ours)      & 5 & 371 & 7 & 31 - 163 & Instruction following  \\
        \midrule
        \multicolumn{6}{c}{\textbf{Static Benchmarks}}\\
        \midrule
        Multilingual MMLU \cite{dac2023okapi}  & 26 &  13062 & N/A & N/A & MC knowledge recall  \\
        Multilingual GSM \cite{shi2023mgsm} & 10  & 250 & N/A & N/A & Math Reasoning \\
        Belebele   \cite{bandarkar-etal-2024-belebele}   & 122 & 900  & N/A & N/A & MC reading comprehension\\
        \bottomrule
    \end{tabular}
    }
    \caption{Descriptive Statistics of Analyzed Benchmarks.}
    \label{tab:dataset_sum}
\end{table*}

\section{Benchmark Datasets and Templates}
We construct multi-lingual \emph{functional} benchmarks by translating (a) the \textbf{100} English GSM-Symbolic templates \citep{gsm-symbolic} and (b) the \textbf{541} English IFEval prompts \citep{zhou2023ifeval} each containing at least one verifiable instruction into French, Spanish, Yoruba, Hindi, and Arabic. Following \citet{bang-etal-2023-multitask, lai-etal-2023-chatgpt}, we refer to languages by ISO 639-1 codes and select them to span resource levels based on CommonCrawl coverage (Table~\ref{tab:lang_class}).

Initial translations are produced with Google Translate \citep{wu2016googlesys}. We then conduct \textbf{native-speaker validation} of all translated GSM-Symbolic templates and all translated IFEval prompts \emph{before} any GSM-Symbolic item instantiation.

\paragraph{Validators and recruitment.}
For each language, we recruited at least \textbf{two native speakers} volunteers to validate translations. Validators were recruited primarily from students studying at a US-based university. For Yoruba, we additionally recruited a native speaker based in Nigeria to ensure dialectal and cultural appropriateness in a low-resource setting.

\paragraph{Quality control, rubric, and prompt filtering.}
Validation followed a structured rubric (Appendix~\ref{sec:translation_validation_rubrics}). Validators compared each translated prompt or template against the English source to ensure fidelity to meaning, intent, and constraint strength, and to verify that instruction realizations remained valid in the target language. Validators discussed and, when possible, corrected problematic translations, and also checked to detect misspellings, duplicated sentences, and template--instance alignment issues. Validators noted that occasional literal word choices remained but did not alter functional constraints or ground-truth mappings.

As part of this validation, they also flagged a subset of original IFEval prompts whose instructions are not meaningfully evaluable in scripts without case distinction (e.g., uppercase/lowercase constraints in Arabic and Hindi). We removed these prompts to preserve cross-lingual comparability and evaluation validity, yielding a \textbf{371}-prompt CL-IFEval set.

After validation, each GSM-Symbolic template was instantiated into \textbf{50 variants per language}, yielding 5{,}000 QA pairs per language, while the validated IFEval prompts constitute CL-IFEval.

For comparison with multi-lingual \emph{static} benchmarks, we evaluate on M-MMLU \citep{dac2023okapi,hendrycks2020measuring}, M-GSM \citep{shi2023mgsm}, and Belebele \citep{bandarkar-etal-2024-belebele}.
The Cross-Lingual GSM Symbolic and Cross-Lingual IFEval datasets are publicly available.
\footnote{\url{https://huggingface.co/datasets/vojewale/Cross-lingualGSMSymbolic}}%
\footnote{\url{https://huggingface.co/datasets/vojewale/Cross-lingualIFEval}}

Further dataset details can be found in Table ~\ref{tab:dataset_sum}. 

\section{Model Evaluation}

For tasks that benefit from step-by-step reasoning, we use Chain-of-Thought (CoT) prompting in the same language as the query where applicable. Unless otherwise stated, \textsc{CL-GSMSym} follows the native 8-shot GSM8K setting, and MGSM uses the \texttt{native\_cot} configuration. All evaluations are orchestrated with the EleutherAI LM Evaluation Harness (\texttt{lm-eval}), which standardizes prompts, decoding, and metrics and is widely used for open-weight models \citep{eval-harness}.

For reproducible consistency in reporting, our evaluations spanned open-source large language models, including instruction-tuned and multilingual variants—Aya 23-35B, Aya Expanse-32B, Gemma-2-9B-it, Qwen3-8B, Mistral-7B-Instruct-v0.3, and Mixtral-8x7B-Instruct-v0.1. Using CL-IFEval, we compute prompt-level strict/loose and instruction-level strict/loose accuracy as in \citet{zhou2023ifeval}. For CL-GSMSym, after template translation, we instantiate \textbf{50 variants per template per language} and evaluate a random sample of 500 items per language under the native 8-shot GSM8K setting.

\begin{figure*}[t]
    \centering
    \resizebox{\textwidth}{!}{
    \includegraphics{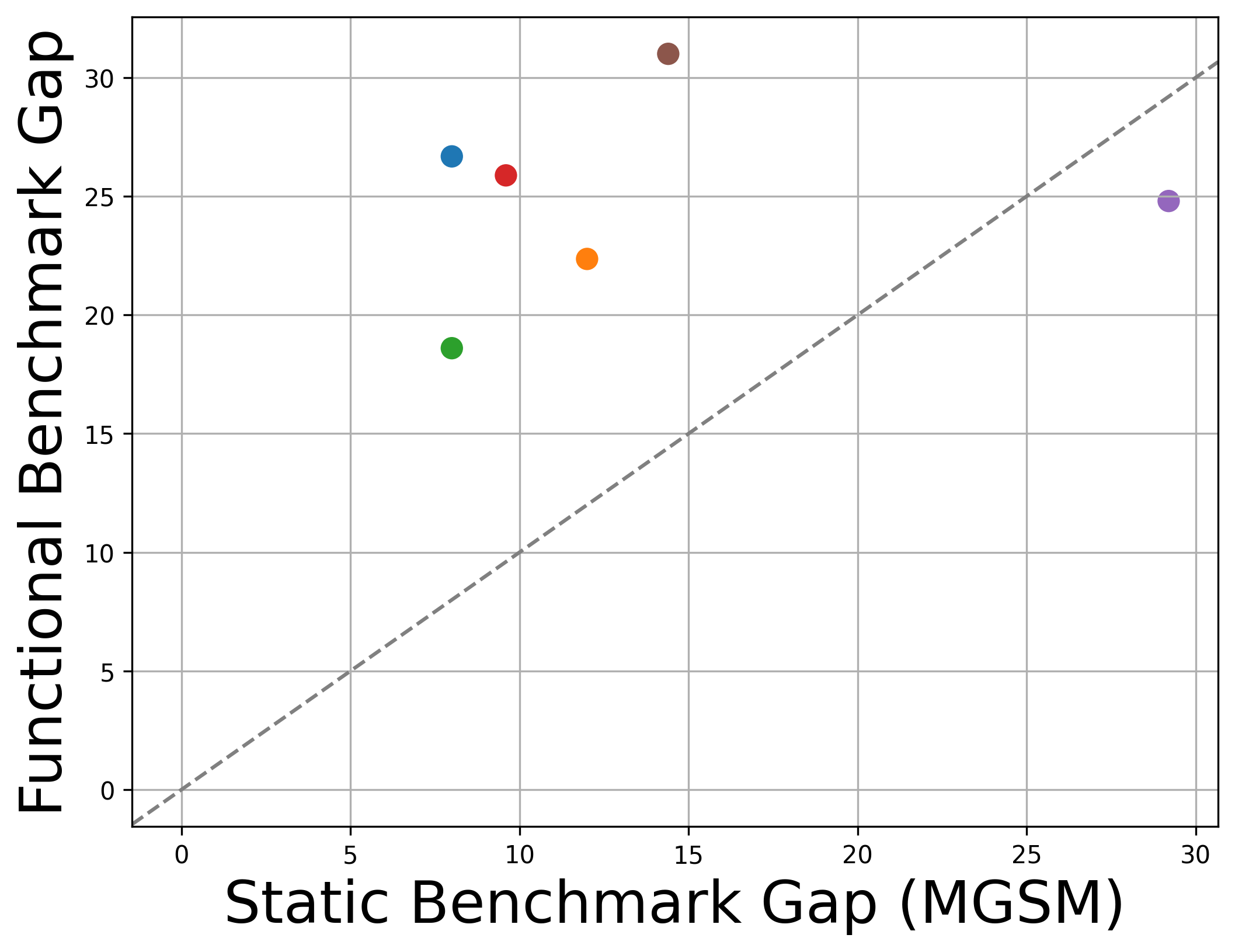}
    \includegraphics{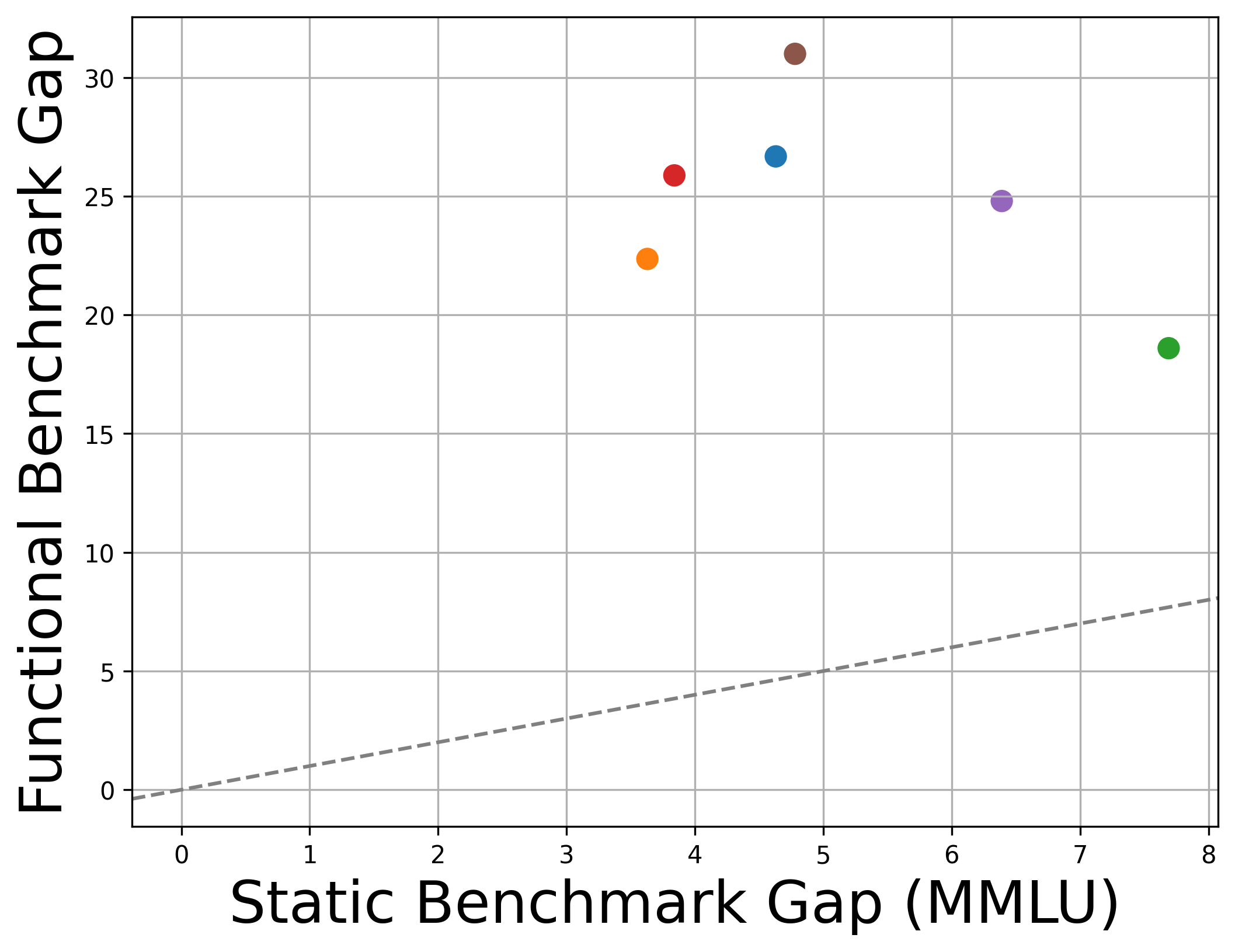}
    \includegraphics{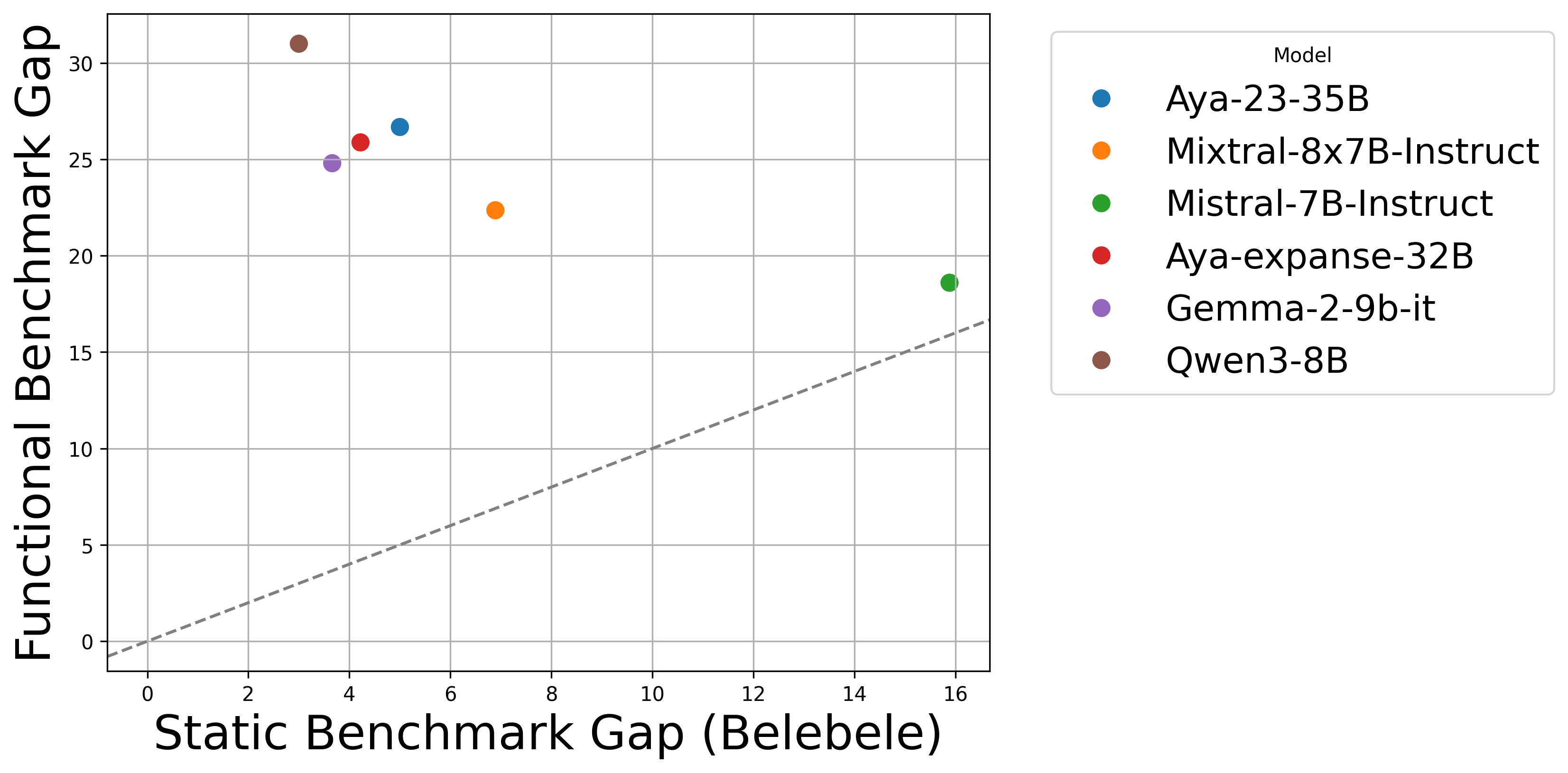}}
    \caption{Correlation plot of Performance Gap between MGSM, MMLU, Belebele (left-to-right) and and CL-IFEval for High Resourced Languages only (en, fr, es). This reveals that measured language performance gaps (i.e. the difference between the performance on the highest performant language and the lowest performant language) are notably larger in functional evaluations than in static data benchmarks.}
    \label{fig:correlation_eval}
\end{figure*}
\section{Results}

\subsection{Average Performance Across Languages \& Model Ranking}

Models perform consistently high for high-resourced languages on static natural language benchmarks (eg. For M-MMLU, English (66\% - 69\%), French (51\% - 65\%) and Spanish (51\% - 65\%) performance is consistently higher; for Belebele, performance across models for high-resource languages is even stronger (79\% – 90\%)). However, there is notable performance variation across models for mathematical reasoning as measured with static data benchmarks -- for example, M-GSM English scores range from 44\% to 86\% across models, while performance in French and Spanish is generally lower, though still relatively competitive and wide-ranging (French: 44\% - 76\%, Spanish: 37\% - 84\%). When available, performance across models decreases on medium resource languages - with scores ranging from 31\% to 48\% in Hindi for M-MMLU, and 41\% to 74\% in Hindi for Belebele. Low-resource languages like Yoruba show drastic performance drops, with scores below 32\% on Belebele. 

\paragraph{Functional benchmarks expose larger within-language variability.} On the other hand, functional benchmarks in natural language, such as CL-IFEval, show a wide range on even English performance (56\% - 88\%), with other high and medium resource languages such as French (36\%- 62\%), Spanish (33\% - 57\%), and Arabic (20\% - 50\%) being similarly wide ranging in performance. CL-GSMSym shows a similar pattern, with a wide range of model performance in high and medium resource languages (English: 49\% - 87\%; French: 39\% - 78\%; Spanish: 34\% - 82\%; Arabic: 18\% - 79\%; Hindi: 15\% - 72\%). Interestingly, in both cases, performance for low resource languages was more consistently low (e.g. in Yoruba, performance on CL-IFEval ranges from 12\% to 23\%, and performance on CL-GSMSym ranges from 4\% to 14\%). 

\paragraph{Model tiering and ranking shifts under functional evaluation.}
For both static benchmarks, the same models are consistently the most performant - typically Qwen3-8B, followed by either Gemma-2-9b-it or Aya-expanse-32B - with the other models not far behind. For functional benchmarks there is a clear discrepancy between a performant class of models (Qwen3-8B, followed by Aya-expanse-32B, followed by Gemma-2-9b-it) and the rest of the models. Unlike with static data benchmarks, this stark performance discrepancy persists, even for low resource languages. For example, on CL-IFEval, the top three models have an English performance range of 70\% to 88\%, whereas the bottom three models all have an English performance range consistently around 60\%. In Yoruba, this performance class discrepancy is still observed, though the performance gap between models is much more narrow. Comparatively, the top three models on the static data benchmark Belebele perform in English at 90\% to 93\% and the bottom three models perform at 80\% to 86\%. Across languages, we can see model rankings change significantly when assessed with static vs functional benchmarking. For example, a typically lower-ranked model like Aya-23-35B, outranks Aya-expanse-32B  and Qwen3-8B in Yoruba on the static Belebele benchmark but not in Yoruba on the functional CL-IFEval benchmark. Similarly, the Aya-23-35B model outranks Gemma-2-9b-it generally on the static M-GSM, even though it consistently performs worse than Gemma-2-9b-it on the functional CL-GSMSym benchmark, for the same languages.

Full details of the performance results for the static data benchmarks and functional benchmarks can be found in Appendix D and E.

\subsection{Language Performance Gap}
We define the language performance gap to be the difference of the model's accuracy  on its lowest performant language, and the highest performant language.
For high- resourced languages, functional evaluations tend to reveal a much larger performance discrepancy than static data benchmarks (see Figure ~\ref{fig:correlation_eval}). Notably, Aya-expanse-32B, a Cohere model marketed specifically for its multi-lingual capability, has a 5.89\% average error gap between high-resourced languages (en, es, fr) across \emph{static} data benchmarks, but a 25.88\% performance gap between those same languages on the \emph{functional} benchmark CL-IFEval. Similarly, Qwen3-8b has fairly low language performance gaps across high resourced languages on static data benchmarks like M-GSM (14.4\%), M-MMLU (4.78\%), and Belebele (3.00\%), but a large language performance gap on the functional CL-IFEval benchmark (31\%). Interestingly, for mathematical reasoning, the opposite can be true -- for example, the language performance gap across high resource languages is lower for the functional CL-GSMSym (e.g. for Qwern3-8b, 8.6\%) than the static data benchmark M-GSM (eg. for Qwern3-8b, 14.4\%). 
Similarly, when including low resource language settings, CL-IFEval tends to show more optimistic numbers (a higher average performance, a lower performance gap) than the static data benchmark, Belebele. In Appendix B, Tables \ref{tab:highresource_results}, \ref{tab:hightomediumresource_results},\ref{tab:hightolowresource_results}, as well as Figure ~\ref{fig:correlation_eval}, we provide further details on these results.

\subsection{Instruction and Template-Level Robustness Across Languages}

\paragraph{Instruction-group breakdowns reveal where cross-lingual generalization fails.}To evaluate instruction-level robustness across languages, we use CL-IFEval to compare model behavior on the IFEval\cite{zhou2023ifeval} default grouped instruction categories. Figures~\ref{fig:robust_clifeval_aya}, \ref{fig:robust_clifeval_gemma}, and \ref{fig:robust_clifeval_qwen} show success rates for Aya-23-35B, Gemma-2-9B, and Qwen3-8B respectively. These plots reveal notable variation in cross-lingual generalization, with Yoruba performing the worst across most instruction groups and showing complete failure in the “Start/End” category for all models.

\paragraph{Template-controlled math variants isolate which problem families drive errors.} To further probe model robustness under controlled prompt variations, we use a subset of CL-GSMSym consisting of 50 samples generated from a fixed set of 10 mathematical question templates from the GSM Symbolic dataset\cite{gsm-symbolic}. Figures~\ref{fig:robust_clgsm_aya}, \ref{fig:robust_clgsm_gemma}, and \ref{fig:robust_clgsm_qwen} display the per-template accuracy of the same three models across languages. Template 3(Appendix \ref{sec:template_examples}), which involves probabilistic inference, consistently produced the lowest performance across models and languages highlighting a persistent weakness in generalizing probabilistic reasoning across linguistic contexts. Interestingly, performance in certain middle-resourced languages (ar, hi) is just as if not more robust for certain types of problems than high resourced languages (e.g. in CL-GSM Symbolic, see Template 2, 4 and 10 results in Figure ~\ref{fig:robust_clgsm_aya}). 

\section{Conclusion}
We introduce two new benchmarks, CL-IFEval and CL-GSMSym, for multi-lingual functional evaluation. Our experiments uncovered major language performance gaps across languages, even for LLMs with robust multilingual claims and strong static data benchmark scores.

\paragraph{Limitations.} We use automated translation tools like Google Translate in constructing CL-IFEval and CL-GSMSym. While these tools offer broad language coverage and facilitate large-scale data generation, they introduce potential inaccuracies, particularly for lower resourced languages like Yoruba, and when dealing with conversions across metric and imperial measurement systems.

Another limitation is the emphasis in our analysis on open-weight models. Our primary focus remains on open-weight models such as Mixtral-8x7B, Mistral-7B, Gemma2-9B-it, Qwen3-8B and AYA models. This creates a possible inherent bias in the scope of our comparisons, as proprietary models may perform significantly better than their open-weight counter-parts. On the other hand, the consistency and transparency of open-weight models make them the preferable object of study -- the incorporation of proprietary models can make results hard to reproduce reliably.  

\paragraph{Future Work.} Future directions include systematically curating higher-quality translations, expanding into multi-lingual code or multi-modal instruction evaluation, and further investigating the robustness and error patterns in both functional and static benchmarks. Also, as functional evaluations involve automatic verification, there is some possibility of extrapolating this framework in the training and evaluation of multi-lingual reasoning models~\cite{yong2025crosslingualreasoningtesttimescaling}.

\section*{Acknowledgments}
The authors would like to thank Zheng-Xin Yong for feedback on the work. This work was supported in part by the MacArthur Foundation, the Mozilla Foundation, and the Heising-Simons Foundation.

\bibliography{custom}

@article{team2024gemma2,
  title={Gemma 2: Improving open language models at a practical size},
  author={Team, Gemma and Riviere, Morgane and Pathak, Shreya and Sessa, Pier Giuseppe and Hardin, Cassidy and Bhupatiraju, Surya and Hussenot, L{\'e}onard and Mesnard, Thomas and Shahriari, Bobak and Ram{\'e}, Alexandre and others},
  journal={arXiv preprint arXiv:2408.00118},
  year={2024}
}

@article{hennigen2023towards,
  title={Towards Verifiable Text Generation with Symbolic References},
  author={Hennigen, Lucas Torroba and Shen, Shannon and Nrusimha, Aniruddha and Gapp, Bernhard and Sontag, David and Kim, Yoon},
  journal={arXiv preprint arXiv:2311.09188},
  year={2023}
}

@article{dussolle2025m,
  title={M-IFEval: Multilingual Instruction-Following Evaluation},
  author={Dussolle, Antoine and D{\'\i}az, Andrea Carde{\~n}a and Sato, Shota and Devine, Peter},
  journal={arXiv preprint arXiv:2502.04688},
  year={2025}
}

@misc{yong2025crosslingualreasoningtesttimescaling,
      title={Crosslingual Reasoning through Test-Time Scaling}, 
      author={Zheng-Xin Yong and M. Farid Adilazuarda and Jonibek Mansurov and Ruochen Zhang and Niklas Muennighoff and Carsten Eickhoff and Genta Indra Winata and Julia Kreutzer and Stephen H. Bach and Alham Fikri Aji},
      year={2025},
      eprint={2505.05408},
      archivePrefix={arXiv},
      primaryClass={cs.CL},
      url={https://arxiv.org/abs/2505.05408}, 
}

@article{ojo2023good,
  title={How good are Large Language Models on African Languages?},
  author={Ojo, Jessica and Ogueji, Kelechi and Stenetorp, Pontus and Adelani, David I},
  journal={arXiv preprint arXiv:2311.07978},
  year={2023}
}

@article{yong2023low,
  title={Low-resource languages jailbreak gpt-4},
  author={Yong, Zheng-Xin and Menghini, Cristina and Bach, Stephen H},
  journal={arXiv preprint arXiv:2310.02446},
  year={2023}
}

@inproceedings{talat2022you,
  title={You reap what you sow: On the challenges of bias evaluation under multilingual settings},
  author={Talat, Zeerak and N{\'e}v{\'e}ol, Aur{\'e}lie and Biderman, Stella and Clinciu, Miruna and Dey, Manan and Longpre, Shayne and Luccioni, Sasha and Masoud, Maraim and Mitchell, Margaret and Radev, Dragomir and others},
  booktitle={Proceedings of BigScience Episode\# 5--Workshop on Challenges \& Perspectives in Creating Large Language Models},
  pages={26--41},
  year={2022}
}

@article{dodge2021documenting,
  title={Documenting large webtext corpora: A case study on the colossal clean crawled corpus},
  author={Dodge, Jesse and Sap, Maarten and Marasovi{\'c}, Ana and Agnew, William and Ilharco, Gabriel and Groeneveld, Dirk and Mitchell, Margaret and Gardner, Matt},
  journal={arXiv preprint arXiv:2104.08758},
  year={2021}
}

@article{zhang2024careful,
  title={A careful examination of large language model performance on grade school arithmetic},
  author={Zhang, Hugh and Da, Jeff and Lee, Dean and Robinson, Vaughn and Wu, Catherine and Song, Will and Zhao, Tiffany and Raja, Pranav and Slack, Dylan and Lyu, Qin and others},
  journal={arXiv preprint arXiv:2405.00332},
  year={2024}
}

@misc{eval-harness,
  author       = {Gao, Leo and Tow, Jonathan and Abbasi, Baber and Biderman, Stella and Black, Sid and DiPofi, Anthony and Foster, Charles and Golding, Laurence and Hsu, Jeffrey and Le Noac'h, Alain and Li, Haonan and McDonell, Kyle and Muennighoff, Niklas and Ociepa, Chris and Phang, Jason and Reynolds, Laria and Schoelkopf, Hailey and Skowron, Aviya and Sutawika, Lintang and Tang, Eric and Thite, Anish and Wang, Ben and Wang, Kevin and Zou, Andy},
  title        = {The Language Model Evaluation Harness},
  month        = 07,
  year         = 2024,
  publisher    = {Zenodo},
  version      = {v0.4.3},
  doi          = {10.5281/zenodo.12608602},
  url          = {https://zenodo.org/records/12608602}
}

@article{singh2024global,
  title={Global mmlu: Understanding and addressing cultural and linguistic biases in multilingual evaluation},
  author={Singh, Shivalika and Romanou, Angelika and Fourrier, Cl{\'e}mentine and Adelani, David I and Ngui, Jian Gang and Vila-Suero, Daniel and Limkonchotiwat, Peerat and Marchisio, Kelly and Leong, Wei Qi and Susanto, Yosephine and others},
  journal={arXiv preprint arXiv:2412.03304},
  year={2024}
}

@article{romanou2024include,
  title={Include: Evaluating multilingual language understanding with regional knowledge},
  author={Romanou, Angelika and Foroutan, Negar and Sotnikova, Anna and Chen, Zeming and Nelaturu, Sree Harsha and Singh, Shivalika and Maheshwary, Rishabh and Altomare, Micol and Haggag, Mohamed A and Amayuelas, Alfonso and others},
  journal={arXiv preprint arXiv:2411.19799},
  year={2024}
}

@misc{zhou2023ifeval,
	title = {Instruction-{Following} {Evaluation} for {Large} {Language} {Models}},
	url = {http://arxiv.org/abs/2311.07911},
	doi = {10.48550/arXiv.2311.07911},
	urldate = {2025-01-28},
	publisher = {arXiv},
	author = {Zhou, Jeffrey and Lu, Tianjian and Mishra, Swaroop and Brahma, Siddhartha and Basu, Sujoy and Luan, Yi and Zhou, Denny and Hou, Le},
	month = nov,
	year = {2023},
	note = {arXiv:2311.07911 [cs]},
}

@article{wu2016googlesys,
  author       = {Yonghui Wu and
                  Mike Schuster and
                  Zhifeng Chen and
                  Quoc V. Le and
                  Mohammad Norouzi and
                  Wolfgang Macherey and
                  Maxim Krikun and
                  Yuan Cao and
                  Qin Gao and
                  Klaus Macherey and
                  Jeff Klingner and
                  Apurva Shah and
                  Melvin Johnson and
                  Xiaobing Liu and
                  Lukasz Kaiser and
                  Stephan Gouws and
                  Yoshikiyo Kato and
                  Taku Kudo and
                  Hideto Kazawa and
                  Keith Stevens and
                  George Kurian and
                  Nishant Patil and
                  Wei Wang and
                  Cliff Young and
                  Jason Smith and
                  Jason Riesa and
                  Alex Rudnick and
                  Oriol Vinyals and
                  Greg Corrado and
                  Macduff Hughes and
                  Jeffrey Dean},
  title        = {Google's Neural Machine Translation System: Bridging the Gap between
                  Human and Machine Translation},
  journal      = {CoRR},
  volume       = {abs/1609.08144},
  year         = {2016},
  url          = {http://arxiv.org/abs/1609.08144},
  eprinttype    = {arXiv},
  eprint       = {1609.08144},
  bibsource    = {dblp computer science bibliography, https://dblp.org}
}

@article{goyal2022flores,
	title = {The {Flores}-101 {Evaluation} {Benchmark} for {Low}-{Resource} and {Multilingual} {Machine} {Translation}},
	volume = {10},
	url = {https://aclanthology.org/2022.tacl-1.30/},
	doi = {10.1162/tacl_a_00474},
	urldate = {2025-01-28},
	journal = {Transactions of the Association for Computational Linguistics},
	author = {Goyal, Naman and Gao, Cynthia and Chaudhary, Vishrav and Chen, Peng-Jen and Wenzek, Guillaume and Ju, Da and Krishnan, Sanjana and Ranzato, Marc'Aurelio and Guzmán, Francisco and Fan, Angela},
	editor = {Roark, Brian and Nenkova, Ani},
	year = {2022},
	note = {Place: Cambridge, MA
Publisher: MIT Press},
	pages = {522--538},
}

@article{hendrycks2020measuring,
  title={Measuring Massive Multitask Language Understanding},
  author={Dan Hendrycks and Collin Burns and Steven Basart and Andy Zou and Mantas Mazeika and Dawn Song and Jacob Steinhardt},
  journal={Proceedings of the International Conference on Learning Representations (ICLR)},
  year={2021}
}

@inproceedings{hasan2021xlsum,
	address = {Online},
	title = {{XL}-{Sum}: {Large}-{Scale} {Multilingual} {Abstractive} {Summarization} for 44 {Languages}},
	shorttitle = {{XL}-{Sum}},
	url = {https://aclanthology.org/2021.findings-acl.413/},
	doi = {10.18653/v1/2021.findings-acl.413},
	urldate = {2025-01-28},
	booktitle = {Findings of the {Association} for {Computational} {Linguistics}: {ACL}-{IJCNLP} 2021},
	publisher = {Association for Computational Linguistics},
	author = {Hasan, Tahmid and Bhattacharjee, Abhik and Islam, Md. Saiful and Mubasshir, Kazi and Li, Yuan-Fang and Kang, Yong-Bin and Rahman, M. Sohel and Shahriyar, Rifat},
	editor = {Zong, Chengqing and Xia, Fei and Li, Wenjie and Navigli, Roberto},
	month = aug,
	year = {2021},
	pages = {4693--4703},
}

@misc{chang2023survey,
	title = {A {Survey} on {Evaluation} of {Large} {Language} {Models}},
	url = {http://arxiv.org/abs/2307.03109},
	doi = {10.48550/arXiv.2307.03109},
	urldate = {2025-01-28},
	publisher = {arXiv},
	author = {Chang, Yupeng and Wang, Xu and Wang, Jindong and Wu, Yuan and Yang, Linyi and Zhu, Kaijie and Chen, Hao and Yi, Xiaoyuan and Wang, Cunxiang and Wang, Yidong and Ye, Wei and Zhang, Yue and Chang, Yi and Yu, Philip S. and Yang, Qiang and Xie, Xing},
	month = dec,
	year = {2023},
	note = {arXiv:2307.03109 [cs]},
}

@misc{aryabumi2024aya,
	title = {Aya 23: {Open} {Weight} {Releases} to {Further} {Multilingual} {Progress}},
	shorttitle = {Aya 23},
	url = {http://arxiv.org/abs/2405.15032},
	doi = {10.48550/arXiv.2405.15032},
	urldate = {2025-01-28},
	publisher = {arXiv},
	author = {Aryabumi, Viraat and Dang, John and Talupuru, Dwarak and Dash, Saurabh and Cairuz, David and Lin, Hangyu and Venkitesh, Bharat and Smith, Madeline and Campos, Jon Ander and Tan, Yi Chern and Marchisio, Kelly and Bartolo, Max and Ruder, Sebastian and Locatelli, Acyr and Kreutzer, Julia and Frosst, Nick and Gomez, Aidan and Blunsom, Phil and Fadaee, Marzieh and Üstün, Ahmet and Hooker, Sara},
	month = may,
	year = {2024},
	note = {arXiv:2405.15032 [cs]},
}

@misc{shi2023mgsm,
	title = {Language {Models} are {Multilingual} {Chain}-of-{Thought} {Reasoners}},
	url = {http://arxiv.org/abs/2210.03057},
	doi = {10.48550/arXiv.2210.03057},
	urldate = {2025-01-28},
	publisher = {arXiv},
	author = {Shi, Freda and Suzgun, Mirac and Freitag, Markus and Wang, Xuezhi and Srivats, Suraj and Vosoughi, Soroush and Chung, Hyung Won and Tay, Yi and Ruder, Sebastian and Zhou, Denny and Das, Dipanjan and Wei, Jason},
	month = oct,
	year = {2022},
	note = {arXiv:2210.03057 [cs]},
}

@inproceedings{bandarkar-etal-2024-belebele,
    title = "The Belebele Benchmark: a Parallel Reading Comprehension Dataset in 122 Language Variants",
    author = "Bandarkar, Lucas  and
      Liang, Davis  and
      Muller, Benjamin  and
      Artetxe, Mikel  and
      Shukla, Satya Narayan  and
      Husa, Donald  and
      Goyal, Naman  and
      Krishnan, Abhinandan  and
      Zettlemoyer, Luke  and
      Khabsa, Madian",
    editor = "Ku, Lun-Wei  and
      Martins, Andre  and
      Srikumar, Vivek",
    booktitle = "Proceedings of the 62nd Annual Meeting of the Association for Computational Linguistics (Volume 1: Long Papers)",
    month = aug,
    year = "2024",
    address = "Bangkok, Thailand",
    publisher = "Association for Computational Linguistics",
    url = "https://aclanthology.org/2024.acl-long.44/",
    doi = "10.18653/v1/2024.acl-long.44",
    pages = "749--775"
}

@misc{jiang2023mistral7b,
      title={Mistral 7B}, 
      author={Albert Q. Jiang and Alexandre Sablayrolles and Arthur Mensch and Chris Bamford and Devendra Singh Chaplot and Diego de las Casas and Florian Bressand and Gianna Lengyel and Guillaume Lample and Lucile Saulnier and Lélio Renard Lavaud and Marie-Anne Lachaux and Pierre Stock and Teven Le Scao and Thibaut Lavril and Thomas Wang and Timothée Lacroix and William El Sayed},
      year={2023},
      eprint={2310.06825},
      archivePrefix={arXiv},
      primaryClass={cs.CL},
      url={https://arxiv.org/abs/2310.06825}, 
}

@misc{jiang2024mixtralexperts,
      title={Mixtral of Experts}, 
      author={Albert Q. Jiang and Alexandre Sablayrolles and Antoine Roux and Arthur Mensch and Blanche Savary and Chris Bamford and Devendra Singh Chaplot and Diego de las Casas and Emma Bou Hanna and Florian Bressand and Gianna Lengyel and Guillaume Bour and Guillaume Lample and Lélio Renard Lavaud and Lucile Saulnier and Marie-Anne Lachaux and Pierre Stock and Sandeep Subramanian and Sophia Yang and Szymon Antoniak and Teven Le Scao and Théophile Gervet and Thibaut Lavril and Thomas Wang and Timothée Lacroix and William El Sayed},
      year={2024},
      eprint={2401.04088},
      archivePrefix={arXiv},
      primaryClass={cs.LG},
      url={https://arxiv.org/abs/2401.04088}, 
}

@misc{cobbe2021training,
    title={Training Verifiers to Solve Math Word Problems},
    author={Karl Cobbe and Vineet Kosaraju and Mohammad Bavarian and Jacob Hilton and Reiichiro Nakano and Christopher Hesse and John Schulman},
    year={2021},
    eprint={2110.14168},
    archivePrefix={arXiv},
    primaryClass={cs.LG}
}

@article{qwen3,
    title={Qwen3 Technical Report}, 
    author={An Yang and Anfeng Li and Baosong Yang and Beichen Zhang and Binyuan Hui and Bo Zheng and Bowen Yu and Chang Gao and Chengen Huang and Chenxu Lv and Chujie Zheng and Dayiheng Liu and Fan Zhou and Fei Huang and Feng Hu and Hao Ge and Haoran Wei and Huan Lin and Jialong Tang and Jian Yang and Jianhong Tu and Jianwei Zhang and Jianxin Yang and Jiaxi Yang and Jing Zhou and Jingren Zhou and Junyang Lin and Kai Dang and Keqin Bao and Kexin Yang and Le Yu and Lianghao Deng and Mei Li and Mingfeng Xue and Mingze Li and Pei Zhang and Peng Wang and Qin Zhu and Rui Men and Ruize Gao and Shixuan Liu and Shuang Luo and Tianhao Li and Tianyi Tang and Wenbiao Yin and Xingzhang Ren and Xinyu Wang and Xinyu Zhang and Xuancheng Ren and Yang Fan and Yang Su and Yichang Zhang and Yinger Zhang and Yu Wan and Yuqiong Liu and Zekun Wang and Zeyu Cui and Zhenru Zhang and Zhipeng Zhou and Zihan Qiu},
    journal = {arXiv preprint arXiv:2505.09388},
    year={2025}
}

@inproceedings{dac2023okapi,
    title = "Okapi: Instruction-tuned Large Language Models in Multiple Languages with Reinforcement Learning from Human Feedback",
    author = "Lai, Viet  and
      Nguyen, Chien  and
      Ngo, Nghia  and
      Nguyen, Thuat  and
      Dernoncourt, Franck  and
      Rossi, Ryan  and
      Nguyen, Thien",
    editor = "Feng, Yansong  and
      Lefever, Els",
    booktitle = "Proceedings of the 2023 Conference on Empirical Methods in Natural Language Processing: System Demonstrations",
    month = dec,
    year = "2023",
    address = "Singapore",
    publisher = "Association for Computational Linguistics",
    url = "https://aclanthology.org/2023.emnlp-demo.28/",
    doi = "10.18653/v1/2023.emnlp-demo.28",
    pages = "318--327"
}

@misc{liu2024mitigatinghallucinationlargemultimodal,
      title={Mitigating Hallucination in Large Multi-Modal Models via Robust Instruction Tuning}, 
      author={Fuxiao Liu and Kevin Lin and Linjie Li and Jianfeng Wang and Yaser Yacoob and Lijuan Wang},
      year={2024},
      eprint={2306.14565},
      archivePrefix={arXiv},
      primaryClass={cs.CV},
      url={https://arxiv.org/abs/2306.14565}, 
}

@misc{dang2024ayaexpansecombiningresearch,
      title={Aya Expanse: Combining Research Breakthroughs for a New Multilingual Frontier}, 
      author={John Dang and Shivalika Singh and Daniel D'souza and Arash Ahmadian and Alejandro Salamanca and Madeline Smith and Aidan Peppin and Sungjin Hong and Manoj Govindassamy and Terrence Zhao and Sandra Kublik and Meor Amer and Viraat Aryabumi and Jon Ander Campos and Yi-Chern Tan and Tom Kocmi and Florian Strub and Nathan Grinsztajn and Yannis Flet-Berliac and Acyr Locatelli and Hangyu Lin and Dwarak Talupuru and Bharat Venkitesh and David Cairuz and Bowen Yang and Tim Chung and Wei-Yin Ko and Sylvie Shang Shi and Amir Shukayev and Sammie Bae and Aleksandra Piktus and Roman Castagné and Felipe Cruz-Salinas and Eddie Kim and Lucas Crawhall-Stein and Adrien Morisot and Sudip Roy and Phil Blunsom and Ivan Zhang and Aidan Gomez and Nick Frosst and Marzieh Fadaee and Beyza Ermis and Ahmet Üstün and Sara Hooker},
      year={2024},
      eprint={2412.04261},
      archivePrefix={arXiv},
      primaryClass={cs.CL},
      url={https://arxiv.org/abs/2412.04261}, 
}

@inproceedings{bang-etal-2023-multitask,
    title = "A Multitask, Multilingual, Multimodal Evaluation of {C}hat{GPT} on Reasoning, Hallucination, and Interactivity",
    author = "Bang, Yejin  and
      Cahyawijaya, Samuel  and
      Lee, Nayeon  and
      Dai, Wenliang  and
      Su, Dan  and
      Wilie, Bryan  and
      Lovenia, Holy  and
      Ji, Ziwei  and
      Yu, Tiezheng  and
      Chung, Willy  and
      Do, Quyet V.  and
      Xu, Yan  and
      Fung, Pascale",
    editor = "Park, Jong C.  and
      Arase, Yuki  and
      Hu, Baotian  and
      Lu, Wei  and
      Wijaya, Derry  and
      Purwarianti, Ayu  and
      Krisnadhi, Adila Alfa",
    booktitle = "Proceedings of the 13th International Joint Conference on Natural Language Processing and the 3rd Conference of the Asia-Pacific Chapter of the Association for Computational Linguistics (Volume 1: Long Papers)",
    month = nov,
    year = "2023",
    address = "Nusa Dua, Bali",
    publisher = "Association for Computational Linguistics",
    url = "https://aclanthology.org/2023.ijcnlp-main.45/",
    doi = "10.18653/v1/2023.ijcnlp-main.45",
    pages = "675--718"
}

@inproceedings{lai-etal-2023-chatgpt,
    title = "{C}hat{GPT} Beyond {E}nglish: Towards a Comprehensive Evaluation of Large Language Models in Multilingual Learning",
    author = "Lai, Viet Dac  and
      Ngo, Nghia  and
      Pouran Ben Veyseh, Amir  and
      Man, Hieu  and
      Dernoncourt, Franck  and
      Bui, Trung  and
      Nguyen, Thien Huu",
    editor = "Bouamor, Houda  and
      Pino, Juan  and
      Bali, Kalika",
    booktitle = "Findings of the Association for Computational Linguistics: EMNLP 2023",
    month = dec,
    year = "2023",
    address = "Singapore",
    publisher = "Association for Computational Linguistics",
    url = "https://aclanthology.org/2023.findings-emnlp.878/",
    doi = "10.18653/v1/2023.findings-emnlp.878",
    pages = "13171--13189"
}

@misc{gsm-symbolic,
title = {GSM-Symbolic: Understanding the Limitations of Mathematical Reasoning in Large Language Models},
author = {Iman Mirzadeh and Keivan Alizadeh and Hooman Shahrokhi and Oncel Tuzel and Samy Bengio and Mehrdad Farajtabar},
year = {2024},
URL = {https://arxiv.org/abs/2410.05229}
}

\newpage
\onecolumn
\appendix

\section{Additional Experimental Details}

\begin{table*}[ht]
    \centering 
    \resizebox{\linewidth}{!}{%
    \begin{tabular}{lccccc}
        \toprule
        Benchmark & Avail. Lang & Prompts/Lang & Templates & Samples/Template & Task \\
        \midrule
        \multicolumn{6}{c}{\textbf{Functional Benchmarks}}\\
        \midrule
        Cross-Lingual GSM Symbolic (Ours)      & 5 & 5000 & 100 & 50 & Math Reasoning  \\
        Cross-Lingual IFEval (Ours)      & 5 & 371 & 7 & 31 - 163 & Instruction following  \\
        \midrule
        \multicolumn{6}{c}{\textbf{Static Benchmarks}}\\
        \midrule
        Multilingual MMLU \cite{dac2023okapi}  & 26 &  13062 & N/A & N/A & MC knowledge recall  \\
        Multilingual GSM \cite{shi2023mgsm} & 10  & 250 & N/A & N/A & Math Reasoning \\
        Belebele   \cite{bandarkar-etal-2024-belebele}   & 122 & 900  & N/A & N/A & MC reading comprehension\\
        \bottomrule
    \end{tabular}%
    }
    \label{tab:mmmlu_results}
    \caption{Descriptive Statistics of Analyzed Benchmarks.}
\end{table*}

\subsection{Models}
Our evaluations spanned open-source large language models, including instruction-tuned and multilingual variants. Below, we provide details on each model:
\begin{itemize}
    \item \textbf{Aya 23-35B}: A multilingual instruction-tuned model based on Cohere's Command framework \citep{aryabumi2024aya}.
    \item \textbf{Aya Expanse-32B}: Part of the Aya Expanse series, designed to enhance multilingual performance capabilities\citep{dang2024ayaexpansecombiningresearch}.
    \item \textbf{Gemma-2-9B-it}: An instruction-tuned model trained on 8 trillion tokens from diverse sources, including web documents, code, and mathematical text. It is optimized for a wide range of text generation tasks, including question answering, summarization, and code understanding \citep{team2024gemma2}.
    \item \textbf{Qwen3-8B}: A dense and mixture-of-experts (MoE) model supporting 100+ languages. It includes specialised modes for logical reasoning, code generation, and agent-based tasks \citep{qwen3}.
    \item \textbf{Mistral-7B-Instruct-v0.3}: An open-source instruction fine-tuned variant of Mistral-7B \citep{jiang2023mistral7b}.
    \item \textbf{Mixtral-8x7B-Instruct-v0.1}: \citep{jiang2024mixtralexperts}A sparse mixture-of-experts model with 12.9B active parameters per token, fine-tuned for instruction tasks.

\end{itemize}

\subsection{Language Resource Classification}
Table \ref{tab:lang_class} classifies the languages used in our evaluation based on their relative resource levels in CommonCrawl.

\begin{table*}[ht]
    \centering
    \begin{tabular}{lcc}
        \toprule
        \textbf{Language} & \textbf{CC-MAIN-2025-05 (\%)} & \textbf{Category} \\
        \midrule
        English (en)  & 43.37  & High-resource (HRL) \\
        French (fr)   & 4.52   & High-resource (HRL) \\
        Spanish (es)  & 4.64   & High-resource (HRL) \\
        Arabic (ar)   & 0.66   & Medium-resource (MRL) \\
        Hindi (hi)    & 0.20   & Medium-resource (MRL) \\
        Yoruba (yo)   & 0.0009 & Extremely low-resource (X-LRL) \\
        \bottomrule
    \end{tabular}
    \caption{Classification of languages based on resource availability in CommonCrawl (CC-MAIN-2025-05).}
    \label{tab:lang_class}
\end{table*}

\section{Full Performance Gap Results}

\subsection{High-Resource Language Performance Gaps}

\begin{table}[H]
    \centering
    \begin{tabular}{lccccc}
        \toprule
        Model & CL-IFEval &  CL-GSMSym &  M-GSM & M-MMLU & Belebele \\
        \midrule
        Aya-23-35B           & 26.68 & 4.4 &  8.0 & 4.63 & 5.00 \\
        Mixtral-8x7B-Instruct & 22.36 & 7.0 &  12.0 & 3.63 & 6.89 \\
        Mistral-7B-Instruct  & 18.60 & 14.0 &  8.0 & 7.69 & 15.88 \\
        Aya-expanse-32B      & 25.88 & 6.2 &  9.6 & 3.84 & 4.22 \\
        Gemma-2-9b-it      & 24.80 & 27.4 & 29.2 & 6.39 & 3.66 \\
        Qwen3-8b      & 31.00 & 8.6 & 14.4 & 4.78 & 3.00 \\
        \bottomrule
    \end{tabular}
    \caption{\textbf{Performance Gap on High-Resource Languages}, as measured by the benchmark accuracy difference between the results on the best performant language and worst performant language across the set of analyzed languages (en, fr, es).}

    \label{tab:highresource_results}

\end{table}

\subsection{High to Medium-Resource Language Performance Gaps}

\begin{table}[H]
    \centering

    \begin{tabular}{lccccc}
        \toprule
        Model & CL-IFEval &  CL-GSMSym &   M-MMLU & Belebele \\
        \midrule
        Aya-23-35B           & 26.68 & 17.6 &  18.08 & 22.45 \\
        Mixtral-8x7B-Instruct & 31.41 & 33.2 & 30.65 & 36.78 \\
        Mistral-7B-Instruct  & 37.74 & 33.8 &  27.92 & 38.44 \\
        Aya-expanse-32B      & 28.03 & 11.0 &  15.29 & 15.45 \\
        Gemma-2-9b-it      & 28.04 & 27.4 & 18.20 & 19.55 \\
        Qwen3-8b      & 37.73 & 27.4 & 20.03 & 23.45 \\
        \bottomrule
    \end{tabular}
    \caption{\textbf{Performance Gap on High to Medium-Resource Languages}, as measured by the benchmark accuracy difference between the results on the best performant language and worst performant language across the set of analyzed languages (en, fr, es, ar, hi).} 
    \label{tab:hightomediumresource_results}
\end{table}

\subsection{High to X Low-Resource Language Gaps}

\begin{table}[H]
    \centering
    \begin{tabular}{lccccc}
        \toprule
        Model & CL-IFEval & CL-GSMSym &  Belebele \\
        \midrule
        Aya-23-35B           & 51.21 & 57.4& 54.12 \\
        Mixtral-8x7B-Instruct & 43.66 & 56.8 & 57.11 \\
        Mistral-7B-Instruct  & 40.98 & 44.2 & 50.11 \\
        Aya-expanse-32B      & 56.87 & 71.6 & 60.45 \\
        Gemma-2-9b-it      & 52.29 & 64.4  & 52.44 \\
        Qwen3-8b      & 69.81 & 73.0 &  62.89 \\
        \bottomrule
    \end{tabular}
    \caption{\textbf{Performance Gap on High to X Low-Resource Languages}, as measured by the benchmark accuracy difference between the results on the best performant language and worst performant language across the set of analyzed languages (en, fr, es, ar, hi, yo).}
    \label{tab:hightolowresource_results}
\end{table}

\section{Static Data Benchmark Results}

\subsection{Multilingual MMLU (M-MMLU)}

The M-MMLU benchmark assesses knowledge and reasoning in a wide range of subjects and languages. The results in Table \ref{tab:mmmlu_results} demonstrate that performance is generally higher in English (0.66--0.69) and other well-resourced languages such as French (0.51--0.65) and Spanish (0.51--0.65). In contrast, performance in medium-resource languages like Hindi is considerably lower, with scores ranging from 0.31 to 0.48. Arabic performance falls between these extremes but is still notably below English and French levels.

\subsection{Multilingual Grade School Math benchmark (MGSM)}

MGSM tests a model's capacity to handle multi-step reasoning problems, particularly in mathematical contexts. The results in Table \ref{tab:mgsm_results} reveal a stark performance gap between English and other languages. English scores range from 0.44 to 0.86, while performance in French and Spanish is generally lower, though still relatively competitive (French: 0.44--0.76, Spanish: 0.37--0.84). French and Spanish scores perform better than seen in M-MMLU, but medium and low-resource languages are not included in this benchmark.

\subsection{Belebele Evaluation}

The Belebele benchmark \cite{bandarkar-etal-2024-belebele} assesses reading comprehension by requiring models to answer multiple-choice questions across a wide variety of languages. Table \ref{tab:belebele_results} highlights significant disparities in performance across resource levels. English comprehension is consistently strong, with scores around 0.85–0.90, while other high-resource languages such as French and Spanish achieve similarly high performance (0.79–0.88). However, low-resource languages like Yoruba show drastic performance drops, with scores below 0.32. Hindi also exhibits relatively weak results (0.41–0.74). 

\section{Full Functional Benchmark Results}
\label{sec:appendix}

\label{sec:appendix:table}

\begin{table}[H]
    \centering

\begin{tabular}{lcccccc}
    \toprule
    Model & English & French & Spanish & Arabic & Hindi & Yoruba \\
    \midrule
    AYA Exp-32B & 74.93 & 55.52 & 49.05 & 46.90 & 49.86 & 18.06 \\
    AYA 23-35B & 67.11 & 48.78 & 40.43 & 43.93 & 40.70 & 15.90 \\
    Mistral-7B-Instruct & 59.57 & 41.78 & 40.97 & 25.33 & 21.83 & 18.59 \\
    Mixtral-8x7B-Instruct & 56.06 & 36.39 & 33.70 & 31.53 & 24.65 & 12.40 \\
    Gemma-2-9B-it & 75.20 & 53.36 & 50.40 & 47.16 & 48.24 & 22.91 \\
    Qwen3-8B & 87.60 & 61.46 & 56.60 & 50.13 & 49.87 & 17.79 \\
    \bottomrule
\end{tabular}

    \label{tab:ifeval_results}
    \caption{Cross-Lingual-IFEval Prompt-level Strict performance across different models and languages.}
\end{table}

\begin{table}[H]
\centering
\small
\begin{tabular}{lcccccc}
\toprule
\textbf{Language} & \textbf{Model} & \textbf{Strict PL (\%)} & \textbf{Strict IL (\%)} & \textbf{Loose PL (\%)} & \textbf{Loose IL (\%)} \\
\midrule

English & AYA 23-35B & 67.11 & 73.75 & 70.89 & 77.41 \\
        & Gemma2-9B-it & 75.20 & 81.08 & 79.24 & 84.17 \\
        & AYA Exp-32B & 74.93 & 79.92 & 78.97 & 83.39 \\
        & Mistral-7B & 59.57 & 68.15 & 64.42 & 72.39 \\
        & Mixtral-8x7B & 56.06 & 64.48 & 61.19 & 69.50 \\
        & Qwen3-8B & 87.60 & 90.70 & 91.37 & 93.63 \\
\midrule

French  & AYA 23-35B & 48.78 & 58.30 & 53.90 & 62.94 \\
        & Gemma2-9B-it & 53.36 & 61.38 & 56.06 & 63.89 \\
        & AYA Exp-32B & 55.52 & 63.12 & 58.22 & 65.83 \\
        & Mistral-7B & 41.78 & 51.35 & 46.90 & 56.18 \\
        & Mixtral-8x7B & 36.39 & 45.37 & 39.62 & 49.03 \\
        & Qwen3-8B & 61.46 & 69.11 & 63.07 & 70.46 \\
\midrule

Spanish & AYA 23-35B & 40.43 & 50.77 & 44.20 & 54.24 \\
        & Gemma2-9B-it & 50.40 & 59.07 & 51.75 & 61.00 \\
        & AYA Exp-32B & 49.05 & 58.49 & 51.21 & 60.62 \\
        & Mistral-7B & 40.97 & 50.96 & 43.66 & 53.86 \\
        & Mixtral-8x7B & 33.70 & 43.47 & 37.20 & 47.00 \\
        & Qwen3-8B & 56.60 & 65.06 & 57.41 & 65.83 \\
\midrule

Yoruba  & AYA 23-35B & 15.90 & 22.77 & 16.17 & 23.74 \\
        & Gemma2-9B-it & 22.91 & 31.08 & 25.88 & 34.75 \\
        & AYA Exp-32B & 18.06 & 28.19 & 29.48 & 30.31 \\
        & Mistral-7B & 18.59 & 28.95 & 20.21 & 31.27 \\
        & Mixtral-8x7B & 12.40 & 20.27 & 15.00 & 23.36 \\
        & Qwen3-8B & 17.79 & 27.22 & 20.21 & 30.50 \\
\midrule

Arabic  & AYA 23-35B & 43.93 & 53.47 & 46.90 & 56.37 \\
        & Gemma2-9B-it & 47.16 & 57.14 & 48.79 & 58.88 \\
        & AYA Exp-32B & 46.90 & 56.76 & 49.06 & 58.69 \\
        & Mistral-7B & 25.33 & 36.10 & 28.57 & 39.38 \\
        & Mixtral-8x7B & 31.53 & 40.73 & 35.85 & 45.95 \\
        & Qwen3-8B & 50.13 & 60.62 & 52.30 & 62.16 \\
\midrule

Hindi   & AYA 23-35B & 40.70 & 50.19 & 44.47 & 54.44 \\
        & Gemma2-9B-it & 48.24 & 56.94 & 50.94 & 59.65 \\
        & AYA Exp-32B & 49.86 & 58.69 & 52.56 & 61.78 \\
        & Mistral-7B & 21.83 & 32.24 & 25.06 & 36.67 \\
        & Mixtral-8x7B & 24.65 & 33.24 & 32.83 & 41.31 \\
        & Qwen3-8B & 49.87 & 58.11 & 52.02 & 60.62 \\
\bottomrule
\end{tabular}

\label{tab:if-comp}
\caption{Comprehensive instruction-following results across multiple models and languages. \textbf{PL} = Prompt-Level Accuracy; \textbf{IL} = Instruction-Level Accuracy.}
\end{table}

\begin{table}[H]
    \centering

    \begin{tabular}{lcccccc}
        \toprule
        Model  & English & French & Spanish & Arabic & Hindi & Yoruba\\
        \midrule
        Aya-23-35B           & 56.00 & 55.00 & 59.40 & 61.20 & 43.60 & 3.80 \\
        Mixtral-8x7B-Instruct  & 56.80 & 55.00 & 62.00 & 32.40 & 28.80 & 5.20 \\
        Mistral-7B-Instruct   & 48.60 & 39.40 & 34.60 & 18.40 & 14.80 & 4.40 \\
        Aya-expanse-32B       & 82.60 & 76.40 & 82.20 & 76.80 & 71.60 & 11.00 \\
        Gemma-2-9b-it         & 77.00 & 49.60 & 54.80 & 55.20 & 57.00 & 12.60 \\
        Qwen3-8B              & 86.80 & 78.20 & 81.80 & 78.80 & 59.40 & 13.80 \\
        \bottomrule
    \end{tabular}
    \label{tab:gsm8ksym_results}
    \caption{CL-GSMSym (8-shot) performance across different models and languages.}
\end{table}

\section{Full Static Data Benchmark Results}
\label{sec:appendix:static_db}

\begin{table}[H]
    \centering

    \begin{tabular}{lcccccc}
        \toprule
        Model & English & French & Spanish & Arabic &Hindi & Yoruba \\
        \midrule
        Aya-23-35B           & 0.8556 & 0.8133 &0.8056 & 0.7667 & 0.6311 & 0.3144 \\
        Mixtral-8x7B-Instruct & 0.8522 & 0.8100 &0.7833 & 0.6344 & 0.4844 & 0.2811 \\
        Mistral-7B-Instruct  & 0.7944 & 0.6711 & 0.6356 & 0.5122 & 0.4100 & 0.2933 \\
        Aya-expanse-32B      &0.8989 & 0.8844 & 0.8567 & 0.8367 & 0.7444 & 0.2944 \\
        Gemma-2-9b-it        & 0.9333 & 0.92   & 0.8967 & 0.8867 & 0.7378 & 0.4089 \\
        Qwen3-8B             & 0.9256 & 0.9089 & 0.8956 & 0.87   & 0.6911 & 0.2967 \\
        \bottomrule
    \end{tabular}
    \caption{Belebele performance across different models and languages.}
    \label{tab:belebele_results}
\end{table}

\begin{table}[H]
    \centering

    \begin{tabular}{lccccc}
        \toprule
        Model  & English & French & Spanish & Arabic & Hindi \\
        \midrule
        Aya-23-35B           & 0.6617 & 0.6168 & 0.6154 & 0.5394 & 0.4809 \\
        Mixtral-8x7B-Instruct  & 0.6854 & 0.6508 & 0.6491 & 0.4142 & 0.3789 \\
        Mistral-7B-Instruct   & 0.5900 & 0.5131 & 0.5181 & 0.3328 & 0.3108 \\
        Aya-expanse-32B       & 0.7396 & 0.7012 & 0.7025 & 0.6267 & 0.5867 \\
        Gemma-2-9b-it         & 0.7149  & 0.6635 & 0.6541  & 0.5565 & 0.5329 \\
        Qwen3-8B              & 0.7528  & 0.7050 & 0.7091  & 0.6072 & 0.5525 \\
        \bottomrule
    \end{tabular}
    \caption{M-MMLU(5-shot) performance across different models and languages.}
    \label{tab:mmmlu_results}
\end{table}

\begin{table}[H]
    \centering

    \begin{tabular}{lccc}
        \toprule
        Model & English & French & Spanish \\
        \midrule
        Aya-23-35B           & 0.668 & 0.588 & 0.600 \\
        Mixtral-8x7B-Instruct & 0.632 & 0.512 & 0.536 \\
        Mistral-7B-Instruct  & 0.448 & 0.444 & 0.368 \\
        Aya-expanse-32B      & 0.856 & 0.760 & 0.840 \\
        Gemma-2-9b-it        & 0.664 & 0.372 & 0.512 \\
        Qwen3-8B             & 0.880 & 0.736 & 0.852 \\
        GPT-4o-mini          & 0.656 & 0.556 & 0.432 \\
        Claude 3.5 Sonnet    & 0.900 & 0.716 & 0.812 \\
        \bottomrule
    \end{tabular}
\caption{MGSM (5-shot) performance across different models and languages. We use questions with answers followed by CoT prompt in the same language \texttt{(native\_cot)} as the dataset and strict match score as the evaluation metric.}
\label{tab:mgsm_results}

\end{table}

\section{Robustness Plots}
\subsection{CL-IFEval Plots}

\begin{figure}[H]
    \resizebox{\columnwidth}{!}{\includegraphics{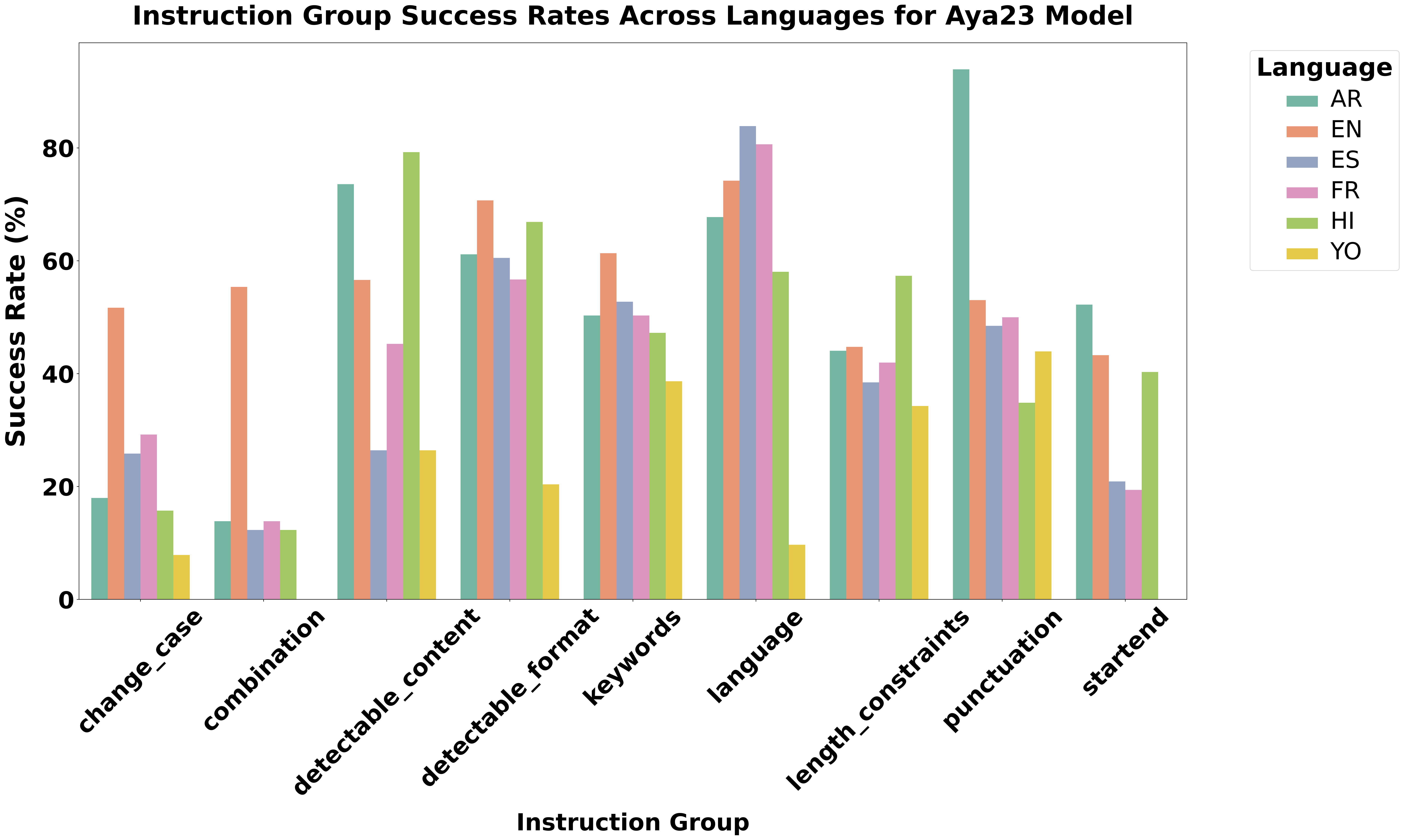}}
    \caption{Cross-Lingual IFEval Aya-23-35B Model Comparison Across Languages}
    \label{fig:robust_clifeval_aya}
\end{figure}

\begin{figure}[H]
    \resizebox{\columnwidth}{!}{\includegraphics{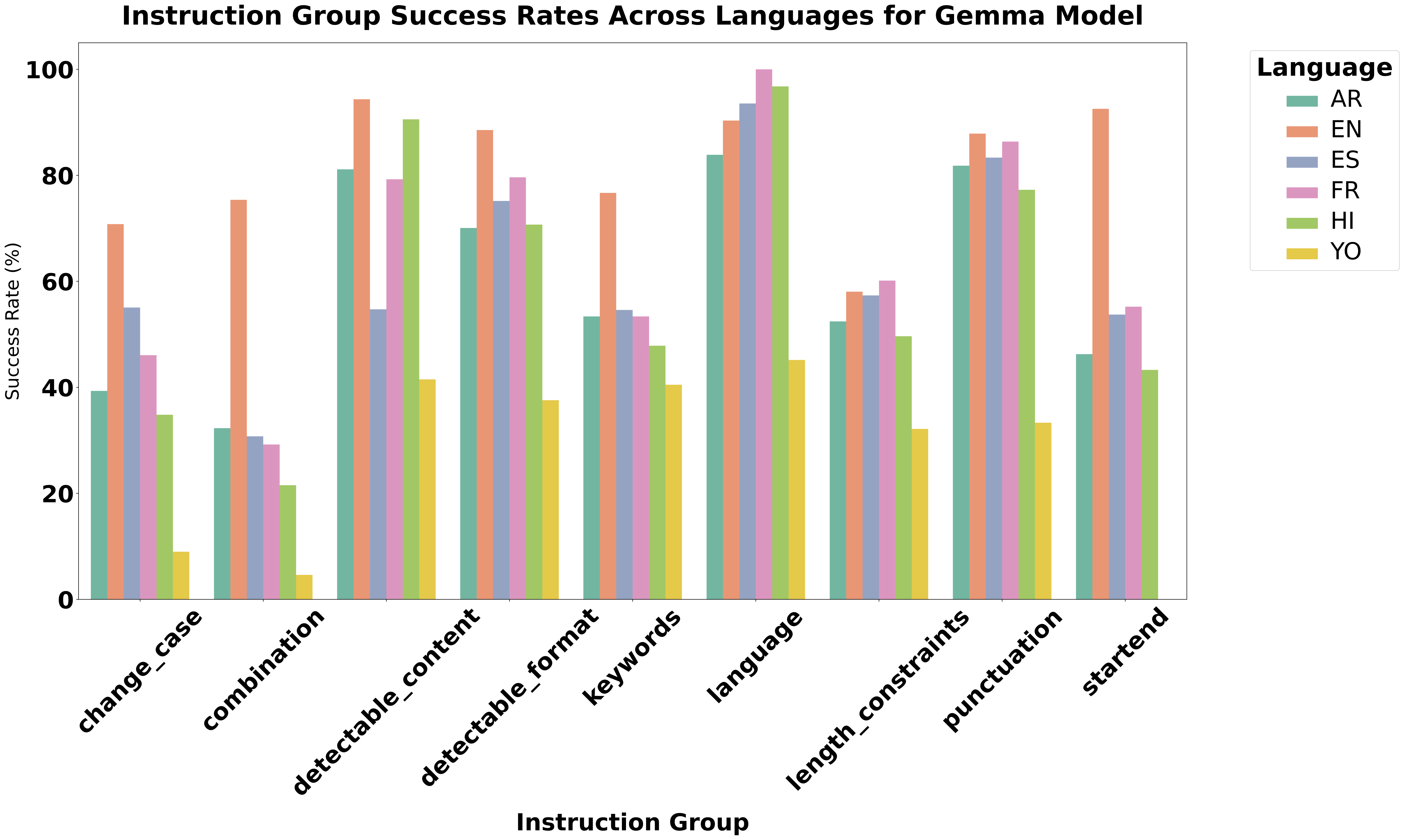}}
    \caption{Cross-Lingual IFEval Gemma-2-9b-it Model Comparison Across Languages}
    \label{fig:robust_clifeval_gemma}
\end{figure}

\begin{figure}[H]
    \resizebox{\columnwidth}{!}{\includegraphics{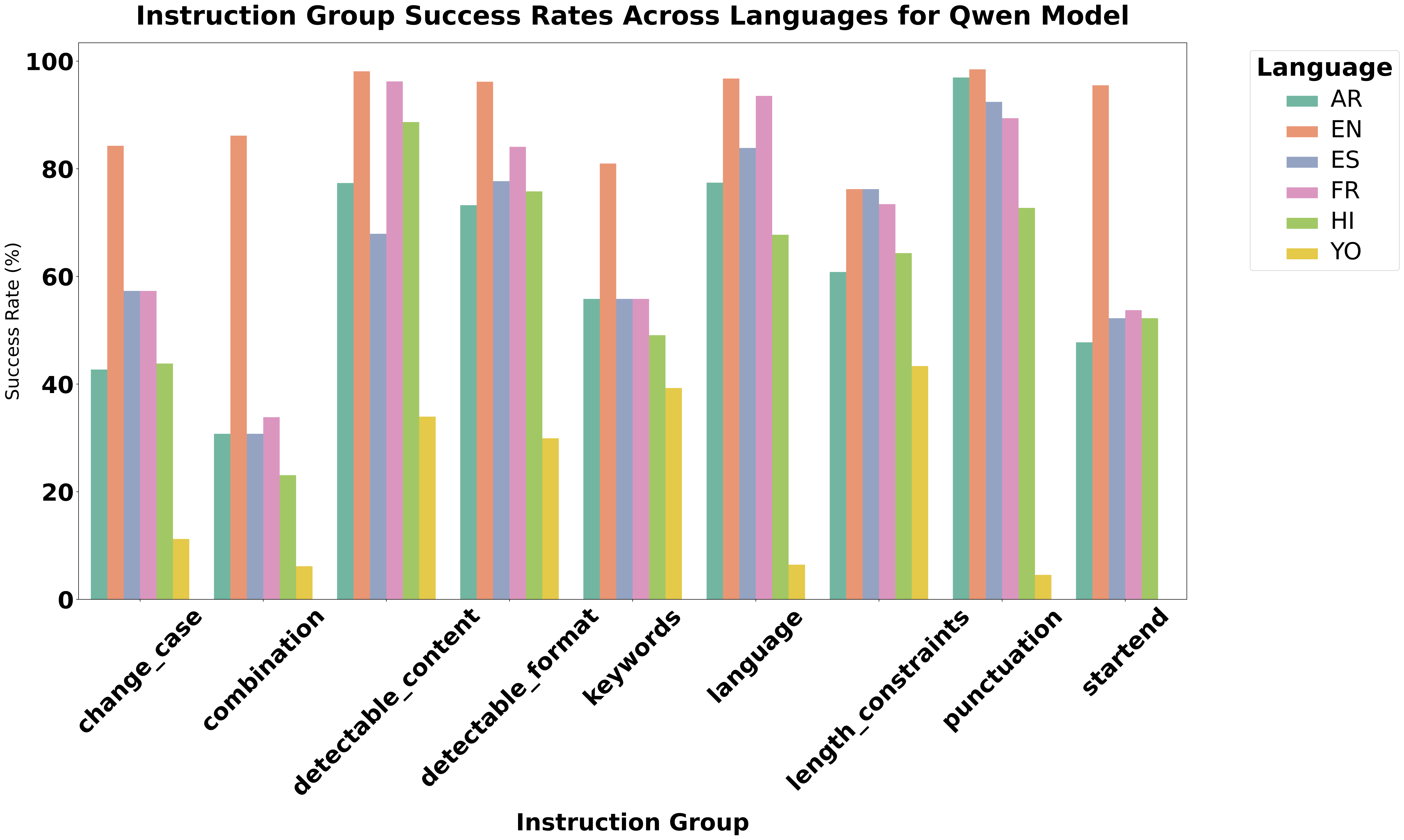}}
    \caption{Cross-Lingual IFEval Qwen3-8b Model Comparison Across Languages}
    \label{fig:robust_clifeval_qwen}
\end{figure}

\subsection{CL-GSMSym Plots}

\begin{figure}[H]
    \resizebox{\columnwidth}{!}{\includegraphics{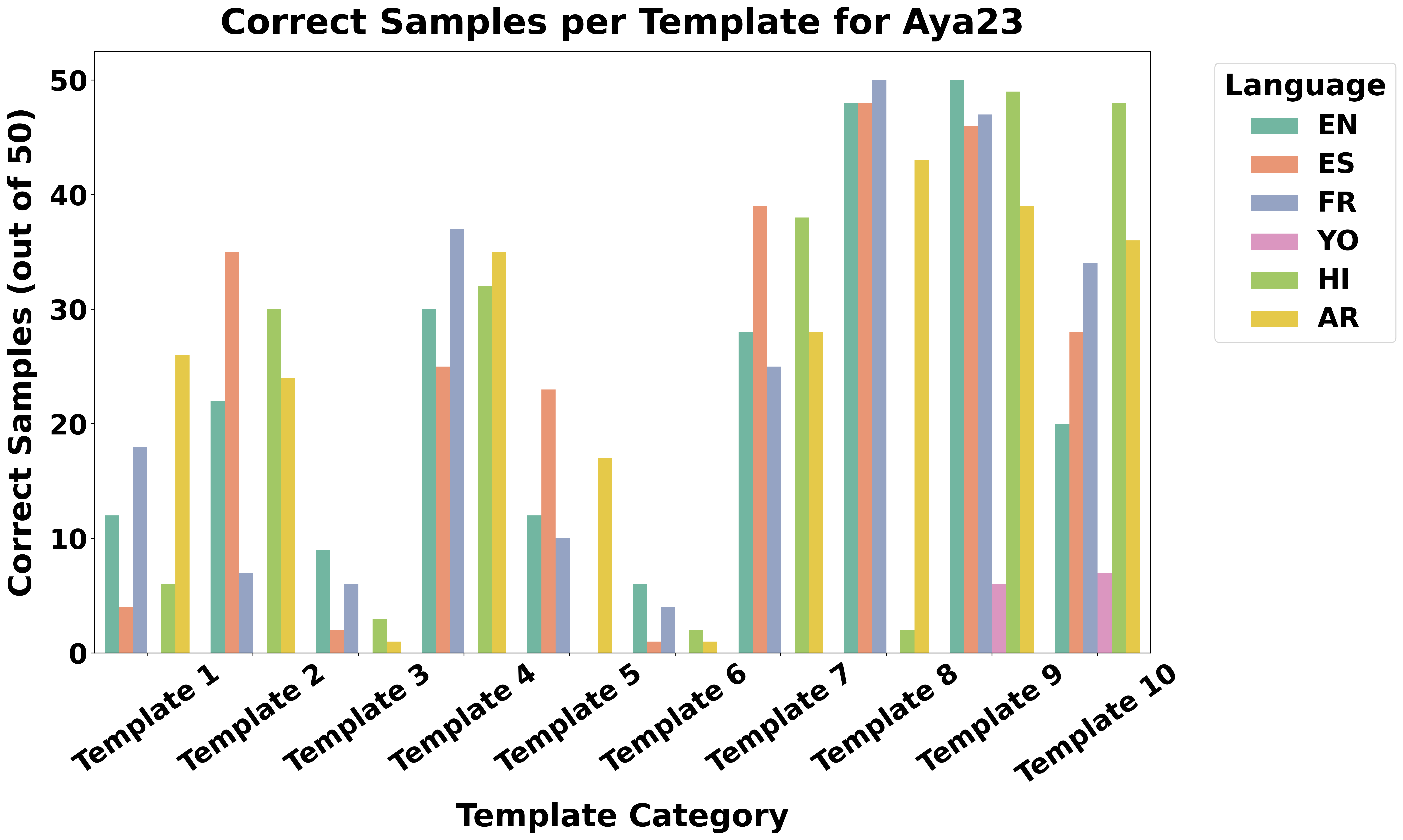}}
    \caption{Cross-Lingual GSMSym Aya-23-35B Template Comparison Across Languages for 50 generated samples of 10 templates}
    \label{fig:robust_clgsm_aya}
\end{figure}

\begin{figure}[H]
    \resizebox{\columnwidth}{!}{\includegraphics{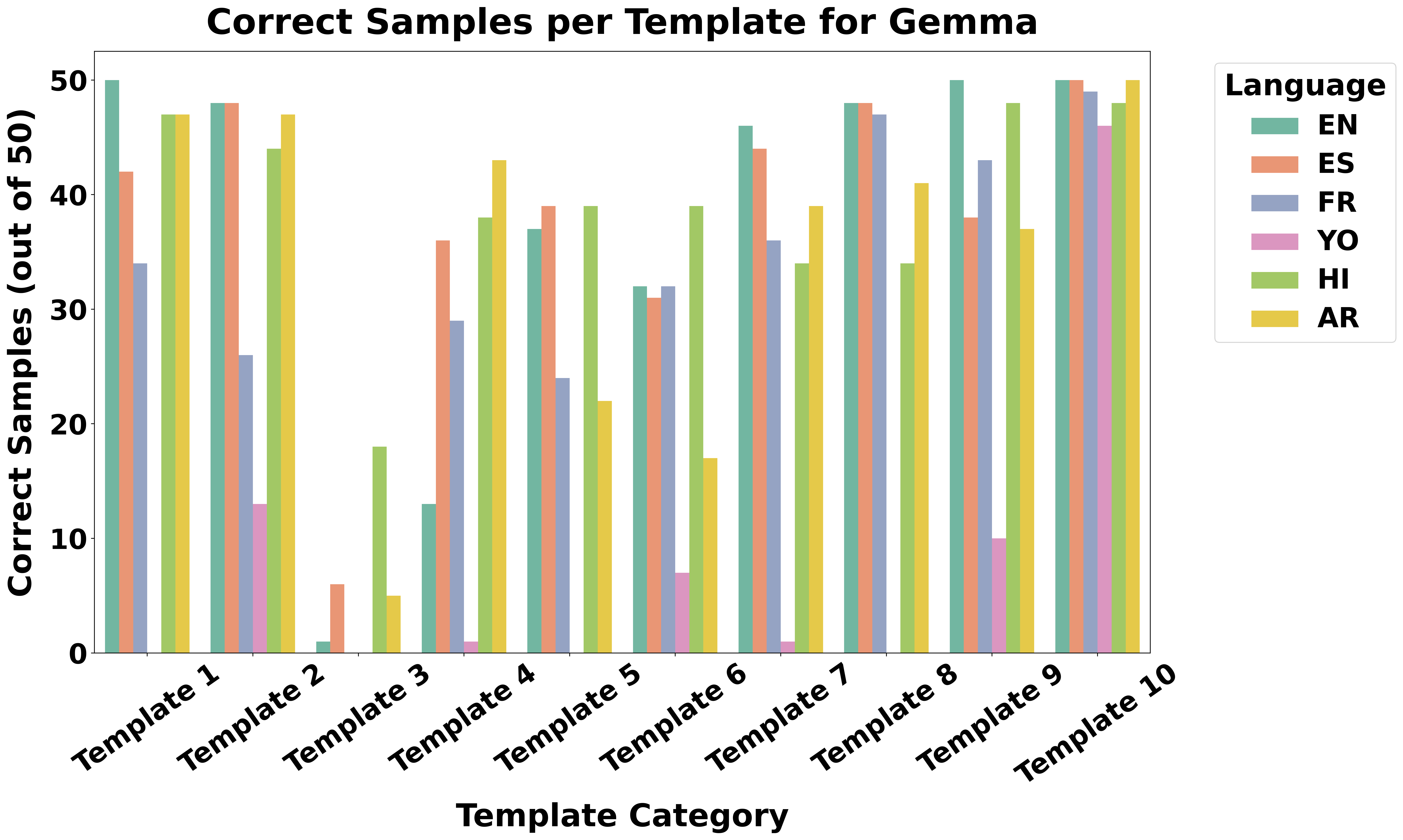}}
    \caption{Cross-Lingual GSMSym Gemma-2-9b-it Template Comparison Across Languages for 50 generated samples of 10 templates}
    \label{fig:robust_clgsm_gemma}
\end{figure}

\begin{figure}[H]
    \resizebox{\columnwidth}{!}{\includegraphics{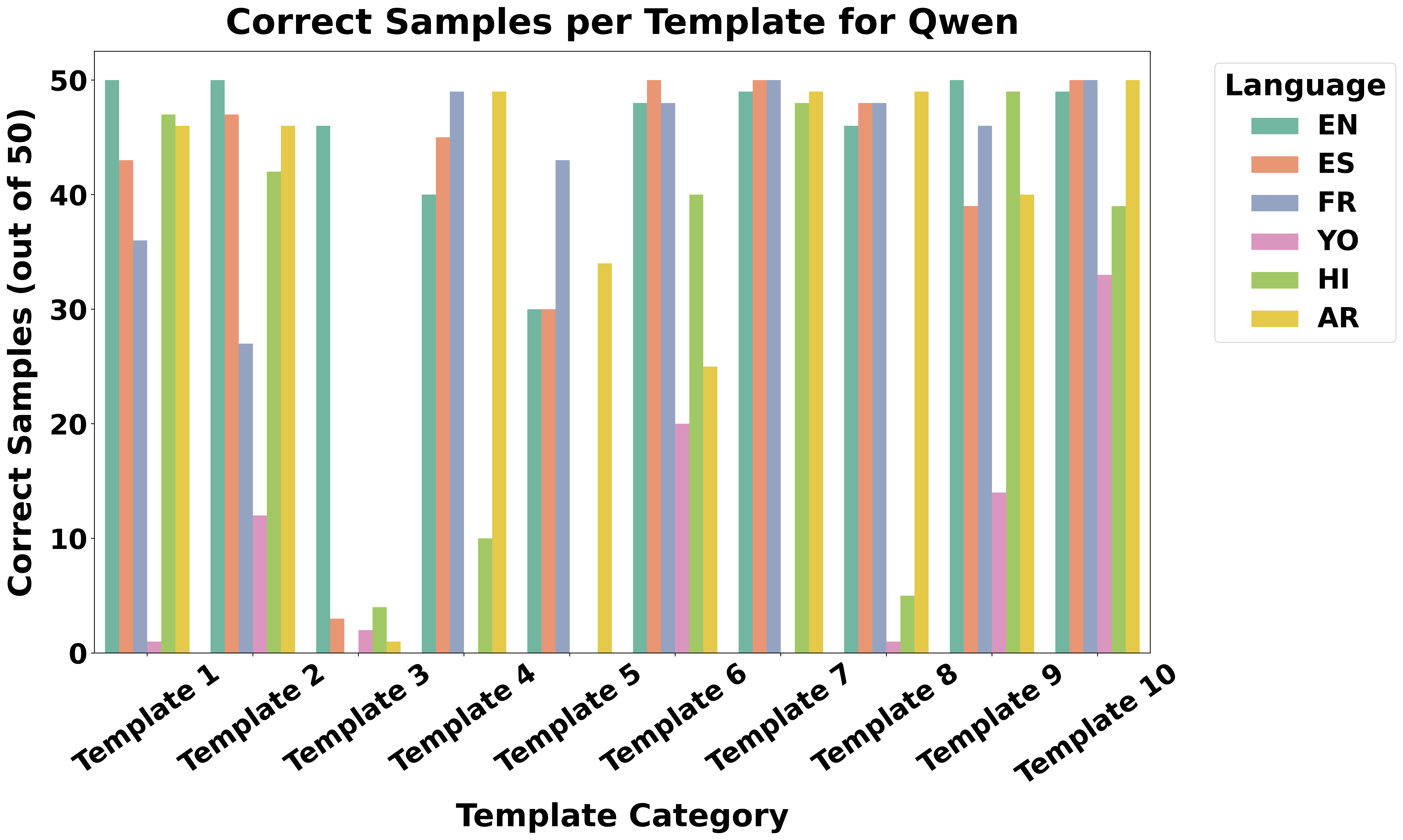}}
    \caption{Cross-Lingual GSMSym Qwen3-8b Template Comparison Across Languages for 50 generated samples of 10 templates}
    \label{fig:robust_clgsm_qwen}
\end{figure}

\section{Translation Validation Rubrics}
\label{sec:translation_validation_rubrics}

\subsection{CL-IFEval Translation Validation (541 Prompts)}
\label{sec:clifeval_translation_validation}

\begin{tcolorbox}[colback=gray!10, colframe=gray, title=1. Objective, boxrule=0.6pt, arc=2mm]
The goal of this validation task is to ensure that every translated prompt accurately represents the intent, structure, and constraints of the original English version while remaining clear and natural in the target language.

Annotators should:
\begin{itemize}
    \item Confirm that the translation faithfully conveys the same meaning and instructions as the English source.
    \item Verify that every element in the instruction\_id\_list is valid and meaningful in the target language.
    \item Flag any instruction IDs that cannot be reasonably expressed or understood in that language.
    \item Correct minor translation errors or omissions directly in the prompt, but only when needed for consistency with the English version.
    \item Record every correction or flag for traceability.
\end{itemize}
\end{tcolorbox}

\begin{tcolorbox}[colback=gray!10, colframe=gray, title=2. Dataset Structure, boxrule=0.6pt, arc=2mm]
Each entry contains:
\begin{itemize}
    \item \texttt{key} -- unique identifier for the instance
    \item \texttt{prompt} -- translated text (the only part that differs from English)
    \item \texttt{instruction\_id\_list} -- the list of instructions applied to the English prompt
    \item \texttt{kwargs} -- parameters or values associated with those instructions
\end{itemize}

Example (simplified):
\begin{verbatim}
{
"key": 1001,
"prompt": "Translated prompt text...",
"instruction_id_list": ["punctuation:no_comma", "length_constraints:number_words"],
"kwargs": [{}, {"num_words": 300}]
}
\end{verbatim}
\end{tcolorbox}

\begin{tcolorbox}[colback=gray!10, colframe=gray, title=3. Validation Criteria, boxrule=0.6pt, arc=2mm]
\begin{itemize}
    \item \textbf{A. Translation Fidelity}
    \begin{itemize}
        \item The translation must convey the same meaning, intent, and constraints as the English prompt.
        \item The tone and phrasing should match the formality and directive style of the English source.
        \item If the translation is missing, ambiguous, or distorts the meaning, revise the prompt to align with the English intent.
        \item When a fix is made, record it with a short note (e.g., ``clarified no-comma rule,'' ``added minimum-count phrase'').
        \item Do not add new content or stylistic embellishments not present in the English prompt.
    \end{itemize}

    \item \textbf{B. Instruction--Prompt Alignment}
    \begin{itemize}
        \item For every instruction in the instruction\_id\_list:
        \begin{itemize}
            \item Realized -- The translated prompt clearly expresses the instruction.
            \item Not Applicable -- The instruction refers to a feature or rule that cannot exist in the target language (e.g., case sensitivity in scripts that lack uppercase/lowercase).
        \end{itemize}
        \item Keep the ID in the metadata, but flag it as ``not applicable in this language.''
        \item Do not remove or modify IDs in the JSON. All flags are recorded separately.
    \end{itemize}

    \item \textbf{C. Linguistic and Logical Consistency}
    \begin{itemize}
        \item The translation must be grammatically correct, culturally neutral, and logically coherent.
        \item Instructions like ``Do not use commas'' or ``Write at least 300 words'' must retain their restrictive force and avoid weaker formulations such as ``Try not to use.''
        \item Quantitative and conditional phrases (e.g., ``at least,'' ``exactly,'' ``no more than'') must match the English structure precisely.
    \end{itemize}

    \item \textbf{D. Formatting and Style}
    \begin{itemize}
        \item Preserve examples, symbols, and URLs exactly as they appear.
        \item Maintain Markdown examples (*highlighted section*, headings, etc.) where relevant.
        \item Ensure punctuation, numbering, and quotation marks follow conventions appropriate for the target language.
        \item Do not introduce additional formatting not found in the original.
    \end{itemize}
\end{itemize}
\end{tcolorbox}

\begin{tcolorbox}[colback=gray!10, colframe=gray, title=4. Correction and Flagging Process, boxrule=0.6pt, arc=2mm]
For each entry:
\begin{enumerate}
    \item Compare the English prompt with the translated version.
    \item Identify whether each instruction ID is not applicable.
    \item If small translation inconsistencies exist, correct them directly in the prompt.
    \item Record every modification or flag, even for minor fixes.
    \item Do not introduce new instructions or stylistic changes that alter the intent.
\end{enumerate}

Permitted corrections include:
\begin{itemize}
    \item Fixing missing or mistranslated phrases that change meaning.
    \item Clarifying numerical constraints or prohibitions (e.g., ``Do not use commas'').
    \item Adjusting phrasing for grammatical correctness.
\end{itemize}

Not permitted:
\begin{itemize}
    \item Adding new examples, stylistic flourishes, or re-interpreting the prompt.
    \item Changing the number of constraints or rewriting the task.
\end{itemize}
\end{tcolorbox}

\begin{tcolorbox}[colback=gray!10, colframe=gray, title=5. Final Review and Consistency Check, boxrule=0.6pt, arc=2mm]
Before final submission:
\begin{itemize}
    \item Ensure that all translations are accurate and natural in their respective languages.
    \item Verify that English-only or script-dependent instruction IDs (e.g., case-based rules) are consistently flagged as ``Not Applicable.''
    \item Confirm that all modifications and flags have been logged.
    \item Export the validated, corrected dataset for each language along with its annotation log in the Log changes column.
\end{itemize}
\end{tcolorbox}

\subsection{CL-GSMSym Translations Validation Guidelines (100 Math Templates)}
\label{sec:clgsmsym_translation_validation}

\begin{tcolorbox}[colback=gray!10, colframe=gray, title=1. Objective, boxrule=0.6pt, arc=2mm]
The goal of this validation task is to ensure that every translated mathematical template accurately preserves the structure, logic, and functionality of the original English version. Annotators should verify that translations remain technically valid, mathematically consistent, and compatible with the generator framework.

The task involves:
\begin{itemize}
    \item Preserving all template markup (\{...\}, \#init:, \#conditions:, \#answer:).
    \item Ensuring the math logic and units match the original.
    \item Correcting small formatting and spacing issues where necessary.
    \item Flagging every fix or inconsistency for audit tracking.
\end{itemize}
\end{tcolorbox}

\begin{tcolorbox}[colback=gray!10, colframe=gray, title=2. Dataset Components, boxrule=0.6pt, arc=2mm]
Each template file contains:
\begin{itemize}
    \item \texttt{question} -- the main text of the math problem.
    \item \texttt{answer} -- the step-by-step explanation and final numeric answer.
    \item \texttt{question\_annotated} -- question with placeholders, variables, and meta blocks.
    \item \texttt{answer\_annotated} -- annotated solution including variables and computed forms.
    \item \texttt{\#init}, \texttt{\#conditions}, \texttt{\#answer} -- meta-sections used by the template generator.
    \item \texttt{id\_orig} / \texttt{id\_shuffled} -- unique identifiers for the template.
\end{itemize}
\end{tcolorbox}

\begin{tcolorbox}[colback=gray!10, colframe=gray, title=3. Validation Criteria, boxrule=0.6pt, arc=2mm]
\begin{itemize}
    \item \textbf{A. Template Markup Preservation}
    \begin{itemize}
        \item Every variable token must remain identical to English:
        
        \texttt{\{var\}}, \texttt{\{var,label\}}, \texttt{\{x,10\}}, \texttt{\{k*y\}}, etc.
        \item The meta-sections (\texttt{\#init:}, \texttt{\#conditions:}, \texttt{\#answer:}) must remain in English and unaltered.
        \item If a translation changed a placeholder (for example, \texttt{\{big\_fish,shark\}} $\rightarrow$ \texttt{\{big\_fish,requin\}}), restore the English version.
        
        Flag: ``Restored placeholder token''.
    \end{itemize}

    \item \textbf{B. Math and Unit Consistency}
    \begin{itemize}
        \item All numeric quantities, operations, and measurement units must match the logic in the English template.
        \item Example: if English uses ``6 inches'', ``10-foot shark'', ``12 inches/foot'', the translation must also use imperial units, not metric.
        \item If the translation converted units (e.g. inches $\rightarrow$ cm, feet $\rightarrow$ meters), revert to the English units.
        
        Flag: ``Unit drift fixed (cm $\rightarrow$ inches)''.
    \end{itemize}

    \item \textbf{C. Angle Brackets and Final Answer}
    \begin{itemize}
        \item Maintain the \texttt{<<...>>} spans exactly as in the English version (these are computation traces).
        \item The final line of every answer must end with:
        
        \texttt{\#\#\#\# <number>} or \texttt{\#\#\#\# \{expression\}}.
        \item If extra spaces, broken symbols, or HTML entities appear, normalize and flag.
        
        Flag: ``Computation span or final answer normalized''.
    \end{itemize}

    \item \textbf{D. Meta Sections (Technical Integrity)}
    \begin{itemize}
        \item The \texttt{\#init:}, \texttt{\#conditions:}, and \texttt{\#answer:} sections must remain in English and in the same order.
        \item If the translation system or annotator translated these lines, restore them verbatim from the original English.
        
        Flag: ``Meta block reverted to EN''.
        \item Ensure indentation and syntax are unchanged --- the generator relies on exact structure.
    \end{itemize}
\end{itemize}
\end{tcolorbox}

\begin{tcolorbox}[colback=gray!10, colframe=gray, title=4. Correction and Flagging Process, boxrule=0.6pt, arc=2mm]
For each template:
\begin{enumerate}
    \item Open both the English and translated versions.
    \item Compare them line by line (especially \texttt{question\_annotated} and \texttt{answer\_annotated}).
    \item Apply small corrections directly:
    \begin{itemize}
        \item Restore missing or broken placeholders.
        \item Fix spacing and punctuation.
        \item Revert units if changed.
    \end{itemize}
    \item Record every modification you make, even small ones.
    \item If the template cannot be fixed to run logically, flag it for exclusion.
\end{enumerate}
\end{tcolorbox}

\begin{tcolorbox}[colback=gray!10, colframe=gray, title=5. Final Review Checklist, boxrule=0.6pt, arc=2mm]
Before final submission, verify that:
\begin{itemize}
    \item All placeholders \texttt{\{...\}} and computation spans \texttt{<<...>>} are intact.
    \item All templates end with the correct final answer format.
    \item Meta sections remain in English, properly indented.
    \item All numbers and units align with the original logic.
    \item Each modification has been recorded in your correction log.
\end{itemize}
\end{tcolorbox}

\begin{tcolorbox}[colback=gray!10, colframe=gray, title=6. Output, boxrule=0.6pt, arc=2mm]
The deliverables per language:
\begin{itemize}
    \item A clean, generator-compatible set of 100 validated templates.
    \item A Column in the google sheet with every correction or flagged issue (for transparency and reproducibility).
\end{itemize}
\end{tcolorbox}

\section{Failure Case Examples for CL-IFEval}

\begin{tcolorbox}[colback=gray!10, colframe=gray, title=Failure Example: \texttt{length\_constraints:number\_words}, width=\textwidth, boxrule=0.5mm, sharp corners, fontupper=\small]
\textbf{Model:} Qwen3-8B \hfill \textbf{Language:} English

\vspace{0.2cm}
\textbf{Prompt:} \textit{Create a riddle about the name Sheldon using only 10 words. Make sure to only use capital letters in your entire response.}

\vspace{0.2cm}
\textbf{Model Response:}  
SHELDON, A NAME WITH A FUNKY RING.

\vspace{0.2cm}
\textbf{Instructions:} 
\begin{itemize}
    \setlength{\itemsep}{0pt}  
    \setlength{\parskip}{0pt}
    \setlength{\topsep}{0pt}
    \setlength{\partopsep}{0pt}
    \item \texttt{change\_case:english\_capital} \faCheck
    \item \texttt{length\_constraints:number\_words} \faTimes
\end{itemize}

\vspace{0.1cm}
\textbf{Failure:} The response contained only 8 words, violating the 10-word constraint.
\end{tcolorbox}

\begin{tcolorbox}[colback=gray!10, colframe=gray, title=Failure Example: \texttt{change\_case:capital\_word\_frequency}, width=\textwidth, boxrule=0.5mm, sharp corners, fontupper=\small]
\textbf{Model:} Qwen3-8B \hfill \textbf{Language:} French

\vspace{0.2cm}
\textbf{Prompt:} \textit{Rédigez un CV pour Matthias Algiers. Utilisez des mots en majuscules pour mettre en avant les compétences clés, mais assurez-vous que les mots entièrement en majuscules apparaissent moins de 10 fois. Enveloppez toute la réponse entre guillemets.}

\vspace{0.2cm}
\textbf{Model Response:}  
"MATTHIAS ALGIERS

RÉSUMÉ PROFESSIONNEL  
Professionnel dévoué et orienté résultats avec une solide expérience en GESTION DE PROJET, PLANIFICATION STRATÉGIQUE et LEADERSHIP D'ÉQUIPE..."

\vspace{0.2cm}
\textbf{Instructions:} 
\begin{itemize}
    \setlength{\itemsep}{0pt}  
    \setlength{\parskip}{0pt}
    \setlength{\topsep}{0pt}
    \setlength{\partopsep}{0pt}
    \item \texttt{change\_case:capital\_word\_frequency} \faTimes
    \item \texttt{startend:quotation} \faCheck
\end{itemize}

\vspace{0.1cm}
\textbf{Failure:} The response included more than 10 fully capitalized words, violating the capital word frequency constraint.
\end{tcolorbox}

\section{Template Examples for CL-GSMSym}
\label{sec:template_examples}
To illustrate the structure and reasoning complexity of items in the \textsc{CL-GSMSym} benchmark, we present representative template-based examples

\begin{tcolorbox}[colback=gray!10, colframe=gray, title=Template 3 Example (French), width=\textwidth, boxrule=0.5mm, sharp corners, fontupper=\small]
\textbf{Language:} French \hfill \textbf{Template:} 3

\vspace{0.2cm}
\textbf{Question:} \textit{Luis lance un dé à quatre faces. Quelle est la probabilité (exprimée en pourcentage) qu'il obtienne un nombre supérieur à 2 plutôt que deux nombres impairs consécutifs ?}

\vspace{0.2cm}
\textbf{Answer:}  
Il y a 2 nombres supérieurs à 2 sur le dé, donc les chances d’en obtenir un sont de \( \frac{2}{4} = 50\% \).  
La probabilité d’obtenir un nombre impair est de 50\%, donc la probabilité d’en obtenir deux d’affilée est \( 0.5 \times 0.5 = 25\% \).  
La différence entre ces deux probabilités est \( 50\% - 25\% = 25\% \).  
\textbf{Final Answer :} \#\#\#\# 25
\end{tcolorbox}

\begin{tcolorbox}[colback=gray!10, colframe=gray, title=Template 3 Example (English), width=\textwidth, boxrule=0.5mm, sharp corners, fontupper=\small]
\textbf{Language:} English \hfill \textbf{Template:} 3

\vspace{0.2cm}
\textbf{Question:} \textit{Faisal is rolling a four-sided die. How much more likely is it (expressed as a percentage) that he rolls a number greater than 1 than that he rolls two even numbers in a row?}

\vspace{0.2cm}
\textbf{Answer:}  
There are 3 numbers greater than 1 on the die, so the chance of rolling one is \( \frac{3}{4} = 75\% \).  
The chance of rolling one even number is 50\%, so the chance of rolling two in a row is \( 0.5 \times 0.5 = 25\% \).  
The difference between these probabilities is \( 75\% - 25\% = 50\% \).  
\textbf{Final Answer:} \#\#\#\# 50
\end{tcolorbox}

\begin{tcolorbox}[colback=gray!10, colframe=gray, title=Template 10 Example (Spanish), width=\textwidth, boxrule=0.5mm, sharp corners, fontupper=\small]
\textbf{Language:} Spanish \hfill \textbf{Template:} 10

\vspace{0.2cm}
\textbf{Question:} \textit{Cuando Emma observa a su primo, saca una variedad de juguetes para él. La bolsa de bloques de construcción tiene 74 bloques dentro. El contenedor de animales de peluche tiene 37 animales de peluche dentro. La torre de anillos apilables tiene 30 anillos multicolores. Emma compró recientemente un tubo de pelotas saltarinas, lo que eleva el número total de juguetes para su primo a 215. ¿Cuántas pelotas saltarinas vinieron en el tubo?}

\vspace{0.2cm}
\textbf{Answer:}  
Sea \( T \) el número de pelotas que rebotan en el tubo.  
Después de comprar el tubo de pelotas, Emma tiene \( 74 + 37 + 30 + T = 141 + T = 215 \) juguetes.  
Por lo tanto, \( T = 215 - 141 = 74 \).  
\textbf{Final Answer:} \#\#\#\# 74
\end{tcolorbox}

\begin{tcolorbox}[colback=gray!10, colframe=gray, title=Template 10 Example (English), width=\textwidth, boxrule=0.5mm, sharp corners, fontupper=\small]
\textbf{Language:} English \hfill \textbf{Template:} 10

\vspace{0.2cm}
\textbf{Question:} \textit{When Winnie watches her nephew, she gets out a variety of toys for him. The bag of building blocks has 72 blocks in it. The bin of stuffed animals has 47 stuffed animals inside. The tower of stacking rings has 29 multicolored rings on it. Winnie recently bought a tube of bouncy balls, bringing her total number of toys for her nephew up to 215. How many bouncy balls came in the tube?}

\vspace{0.2cm}
\textbf{Answer:}  
Let \( T \) be the number of bouncy balls.  
Total number of toys before the balls: \( 72 + 47 + 29 = 148 \).  
So \( 148 + T = 215 \Rightarrow T = 215 - 148 = 67 \).  
\textbf{Final Answer:} \#\#\#\# 67
\end{tcolorbox}

\end{document}